\documentclass[twoside,10.5pt]{article}
\usepackage{jmlr2e}

\usepackage[table]{xcolor}
\definecolor{oxfordblue}{rgb}{0.0, 0.13, 0.28}
\definecolor{linkblue}{rgb}{0.0, 0.20, 0.40}
\usepackage{hyperref} 
\hypersetup{
    colorlinks = true,
    linkcolor = linkblue,
    anchorcolor = linkblue,
    citecolor = linkblue,
    filecolor = linkblue,
    urlcolor = linkblue
}

\usepackage[most]{tcolorbox}
\usepackage{etoolbox} 
\definecolor{faintgray}{RGB}{245,245,245}     
\definecolor{faintborder}{RGB}{230,230,230}   
\definecolor{lightblack}{gray}{0.4}           

\newtcolorbox{question}[1][]{
  enhanced,
  colback=faintgray,
  colframe=faintborder,
  boxrule=0.2pt,
  arc=2mm,
  title=\textcolor{lightblack}{\textbf{Question}},
  fonttitle=\bfseries,
  before upper={\centering\itshape},
  after title={\vspace{0.5ex}},
  boxsep=4pt,
  left=6pt,
  right=6pt,
  top=4pt,
  bottom=4pt,
  #1
}

\usepackage[top=3cm, bottom=3cm, left=2.5cm, right=2.5cm]{geometry}

\usepackage[shortlabels]{enumitem}
\usepackage{amsmath, amssymb, mathabx, comment, mathrsfs}
\usepackage{graphicx}
\usepackage{subcaption}
\usepackage{array, multirow}
\usepackage[font=small,labelfont=bf]{caption} 
\usepackage{lipsum}
\usepackage{amsfonts}
\usepackage{tabularx}
\usepackage{booktabs}
\usepackage{graphicx}
\usepackage{epstopdf}
\usepackage{algorithmic}
\usepackage{mathtools}
\usepackage{tikz}
\usepackage{todonotes}
\usepackage{float}
\definecolor{darkgreen}{rgb}{0, .5, 0}

\ifpdf
\DeclareGraphicsExtensions{.eps,.pdf,.png,.jpg}
\else
\DeclareGraphicsExtensions{.eps}
\fi

\AtBeginDocument{%
\setlength{\oddsidemargin}{\dimexpr(\paperwidth-\textwidth)/2-1in}%
\setlength{\evensidemargin}{\oddsidemargin}%
\setlength{\topmargin}{%
\dimexpr(\paperheight-\textheight)/2-\headheight-\headsep-1in}%
}

\usepackage{lastpage}

\newcommand{\R}{\mathbb{R}}
\newcommand{\N}{\mathbb{N}}
\newcommand{\zod}{[0,1]^d}
\newcommand{\zd}{(0,1)^d}

\newcommand{\id}{\mathrm{id}}
\newcommand{\inter}{\text{int}}

\newcommand{\NODE}{\mathrm{NODE}}

\renewcommand{\N}{\mathbb{N}}
\renewcommand{\R}{\mathbb{R}}
\newcommand{\Rd}{\mathbb{R}^d}

\NewDocumentCommand{\Fl}{o o}{%
    \mathrm{Fl}{\IfValueT{#1}{_{#1}}\IfValueT{#2}{^{#2}}}
}
\NewDocumentCommand{\Diff}{o o o}{%
    \ensuremath{%
        \operatorname{Diff}%
        \IfValueT{#3}{^{#3}}%
        \! \IfValueT{#1}{\left(#1\IfValueT{#2}{,#2}\right)_0%
        }%
    }%
}
\NewDocumentCommand{\Diffk}{o o}{%
    \ensuremath{%
        \operatorname{Diff}%
        \IfValueT{#2}{^{#2}}%
        \! \IfValueT{#1}{\left(#1\right)_0%
        }%
    }%
}
\NewDocumentCommand{\Homeo}{o o o}{%
    \ensuremath{%
        \mathcal{H}%
        \IfValueT{#3}{^{#3}}%
        \IfValueT{#1}{%
            \left(#1\IfValueT{#2}{,#2}\right)_0%
        }%
    }%
}
\NewDocumentCommand{\HomeoI}{o o}{%
    \ensuremath{%
        \mathcal{H}%
        \IfValueT{#2}{^{#2}}%
        \IfValueT{#1}{%
            \left(#1\right)_0%
        }%
    }%
}
\NewDocumentCommand{\Flow}{o}{%
    \operatorname{Flow}{\IfValueT{#1}{\left({#1}\right)}}
}
\newcommand{\eqdef}{\ensuremath{\stackrel{\mbox{\upshape\tiny def.}}{=}}}

\newcommand{\calX}{\mathcal{X}}

\usepackage{amssymb}  
\usepackage{pifont}   
\definecolor{darkred}{RGB}{139,0,0}
\definecolor{darkgreen}{RGB}{0,100,0}

\usepackage{cleveref}

\crefname{corollary}{Corollary}{Corollary}
\crefname{theorem}{Theorem}{Theorem}
\crefname{lemma}{Lemma}{Lemmata}
\crefname{assumption}{assumption}{assumptions}
\Crefname{assumption}{Assumption}{Assumptions}
\crefname{example}{Example}{Examples}
\crefname{proposition}{Proposition}{Proposition}

\makeatletter
\providecommand{\@editor}{}
\makeatother

\begin{document}

\title{Autonomous-Flow-Based Generation}

\author{%
    \name Hossein Rouhvarzi
	\email rouhvarzih@gmail.com\\
    \addr McMaster University and the Vector Institute\\
	Department of Mathematics\\
    1280 Main Street West, Hamilton, Ontario, L8S 4K1, Canada
    \AND
    \name Anastasis Kratsios\thanks{Corresponding author.}
	\email kratsioa@mcmaster.ca\\
    \addr McMaster University and the Vector Institute\\
	Department of Mathematics\\
    1280 Main Street West, Hamilton, Ontario, L8S 4K1, Canada
	}
\maketitle

\begin{abstract}
We show that using autonomous-flow-based generation, one can universally approximate orientation-preserving diffeomorphisms defined on the cube by Neural ODEs with rate $\mathcal{O}(P^{-1/d})$ with $P$ parameters. On the other hand, we show that by using only a single autonomous flow, the class of Neural ODEs is nowhere dense on the cube in dimension $d \ge 2$ .
Under a compact-support$_\id$ condition on $(0,1)^d$, 
we show that using autonomous-flow-based generation, one can universally approximate compactly supported$_\id$ diffeomorphisms on $(0,1)^d$ for any dimension with rate $\mathcal{O}((\frac{P}{\log P})^{-2/d})$ with $P$ parameters and for compactly supported$_\id$ homeomorphisms on $(0,1)^d$ in dimension $d \geq 5$ with rate $\mathcal{O}(P^{-1/(d+1)})$ with $P$ parameters and by a composition of at most $I_d$ autonomous Neural ODEs with the same support$_\id$, where $I_d$ depends only on the dimension.
Moreover, we show that the class of single autonomous flows compactly supported$_\id$ on $(0,1)^d$ is meagre in the space of compactly supported$_\id$ homeomorphisms on $(0,1)^d$ for $d\ge 2$. By linearly lifting the domain into one higher dimension, we obtain a universal approximation result for Lipschitz functions compactly supported on $(0,1)^d$ with rate $\mathcal{O}(P^{-1/(d+1)})$ with $P$ parameters.

\end{abstract}

\begin{keywords}
Autonomous Flow; Neural ODEs; Universal Approximation; Dynamical Systems; Homeomorphism and Diffeomorphism Groups; Structure-Aware AI.
\end{keywords}

\,\,\textbf{MSC (2020):} 41A30, 41A65, 58D05, 58D15, 37C10.

\tableofcontents

\section{Introduction}
\label{s:Introduction}

Autonomous-flow-based generation has revolutionized areas including computational anatomy and image registration~\cite{thottupattu2020method,bostelmann2026stationary}, manifold learning~\cite{hanson2025fitting,vigouroux2026discovering}, kinematic calibration in robotics~\cite{lafferriere1991motion,park1994kinematic,wang2012screw}, ResNets and Neural ODEs~\cite{scagliotti2023deep,KuehnWohrer2026,ZhangGaoUntermanArodz2020,Honda2026}.

One can consider flows as solution of Cauchy ODE systems of vector fields (for a standard definition of flow refer to for example~\cite[Chapter 9]{Lee_2013__SmoothManifolds}). As an example, consider the solution of the Cauchy ODE system below:
\begin{equation}
\label{eq:autonomous_ODE}
    \frac{d x_t^x}{d t}  = V(x_t^x), \quad x_0^x = x,
\end{equation}
where $V :\mathbb{R}^d \to \mathbb{R}^d$ is at least Lipschitz. By the Picard–Lindelöf theorem it has a unique solution. We consider the flow at a fixed time, which we take without loss of generality, to be time-one.
The resulting time-one transformation $\varphi \eqdef x_1^x$ is then \textit{Lipschitz continuous}.
This transformation has a Lipschitz inverse $\varphi^{-1}$ with infinitesimal generator $-V$.

An \textit{autonomous flow} refers to a flow having a time-independent infinitesimal generator. There have been studies to show that the set of autonomous flows is nowhere dense or meagre. 
For instance, for the class of homeomorphisms in the identity component\footnote{Homeomorphisms that are isotopic to identity.} of a compact smooth manifold, in dimension $d\ge2$, the set of homeomorphisms that can be written as a time-one continuous autonomous flow, is proved to be nowhere dense \cite{BonomoVarandas2020}.
Moreover, it is proved for the class of $C^1$-diffeomorphisms on closed connected smooth manifolds \cite{BonattiCrovisierWilkinson2009} that the set of flowable $C^1$-diffeomorphisms is meagre.\footnote{As flows don't have trivial centralizer and the set of diffeomorphisms on that space is completely metrizable so it is a Baire space~\cite[Corollary 25.4]{willard2004general}.} For the classical compact manifold see \cite{Palis1974}.
Unfortunately, for $C^\infty$diffeomorphisms, only a subset of them has been proved to be meagre. Examples include those that satisfy Axiom A and the transversality condition for $d\ge2$~\cite{PalisYoccoz1989}, or Axiom A diffeomorphisms
with the no cycles property that are non-Anosov for $d\ge3$~\cite{Fisher2008}. The full result has been proved only for the circle~\cite{ChernSmale1970}. In this paper we use these facts to analyze the limitations of using single autonomous flows. Namely, we show that single Neural ODEs are nowhere dense when considering the cube, and meagre when considering the compact support$_\id$ case.

On the other hand, it is well known that one can represent diffeomorphisms as a composition of time-one flows~\cite{GrongSchmeding2025,Luckaki1978RepDiff,ishikawa2023universal,Agrachev2009}.
In this paper we prove it again for compactly supported$_\id$ diffeomorphisms on $\zd$ and show that, by considering this representation, one can have universality for Neural ODEs. Moreover, in this support$_\id$, we show that the minimal number of time-one autonomous flows which is needed to represent is bounded by a constant $I_d$ depending only on the dimension $d$. Furthermore, we use these facts to approximate compactly supported$_\id$ homeomorphisms on $\zd$ when $d \geq 5$ using composition of time-one autonomous flows thereby obtaining a universal approximation by Neural ODEs. Unfortunately, our approach doesn't extend to the homeomorphisms defined on the cube, as we will show that diffeomorphisms on the cube are not dense in the space of homeomorphisms defined on the cube. Although we will show one can quantitatively approximate diffeomorphisms on the cube, we cannot prove whether there exists any similar constant $I_d$.

Traditionally, in approximation-theoretic deep learning theorems, they consider the cube $\zod$ as the domain of the maps~\cite{mhaskar2016deep,yarotsky2017error,petersen2018optimal,elbrachter2021deep,kratsios2022do,zhang2022deep,siegel2023optimal}. This is because analytically, as it is compact and bounded, norms and approximation rates are controlled, and extension and rescaling arguments are available. Moreover, geometrically and computationally, its product structure makes grids, ReLU constructions, triangulations, and coordinate-wise arguments easy. In our paper, to study functions on a compactly supported$_\id$ domain defined on $\Rd$, because of technical topological and analytical convenience, (e.g. to use a ReLU bump function, BIP-meagreness etc.) we will consider the quite similar domain $\zd$. This means, we will study maps $\varphi$ that are  compactly supported$_\id$ on $\zd$; This is also compatible with applications considering group of diffeomorphisms~\cite{Dpuis1998VarIM} where they consider open bounded subset of $\Rd$.

Note that by Alexander's trick~\cite{AlexandersTrick} every such homeomorphism must be orientation-preserving.
We denote the sets of compactly supported$_\id$ orientation-preserving homeomorphisms and diffeomorphisms on $\zd$ by $\Homeo[\Rd][\zd]$ and $\Diff[\Rd][\zd][\infty]$ respectively. Note that these are not vector spaces. It is known that $\Diff[\Rd][\zd][\infty]$ is a topological $l^2 \times \R^\infty$-manifold equipped with the Whitney topology. This is only true for $\Homeo[\Rd][\zd]$ when $d=2$~\cite{BanakhMineSakaiYagasaki2011}. Note that $\R^\infty$ is a non-metrizable LF-space~\cite{BanakhRepovs2012}, which places it outside the scope of the available constructive approximation toolbox~\cite{Pinkus1985,LorentzGolitschekMakovoz1996,cohen2022optimal,petrova2023Lipschitz}, where they assume a Banach space structure. On the other hand $\Diff[\Rd][\zd][\infty]$ and $\Homeo[\Rd][\zd]$ possess a rich and well-studied infinite-dimensional topological group structure~\cite{DimHomeo2001Brechner,mather1974commutators,mather1984curious,thurston1974foliations,fathi1980structure,Banyaga_1997__DiffeoBook,haller2013smooth,fukui2019uniform}.  
Moreover, unlike $\Diff[\Rd][\zd][\infty]$ which is a smooth regular infinite-dimensional Lie group~\cite{Michor2016Manifolds,Michor1980} and the global exponential map exists (in the sense of infinite-dimensional Lie groups; cf.~\cite{Kriegl1997,BauerHarmsMichor2025}), for $\Homeo[\Rd][\zd]$ we do not have access to a global exponential map to transfer results from linear spaces onto $\Homeo[\Rd][\zd]$, (as in~\cite{kratsios2020non,kratsios2022universalpapon}), to apply signature-based machine learning methods, ~\cite{cass2024lecturenotes,bayer2025optimalstopping}, rooted in rough path theory~\cite{lyons1998differential,gubinelli2004controlling,HamblyLyons2010}.
In short, there are currently no available tools to apply classical approximation results from linear spaces onto $\Homeo[\Rd][\zd]$ using a global exponential map. We are required to develop new approximation-theoretic techniques suited to its non-vectorial structure, where our approximation, or hypothesis, class consists only of (neural network-based) homeomorphisms in $\Homeo[\Rd][\zd]$. 

We note some \(L^p\) versions of these results are known for finite $p$ via controllability arguments~\cite{cuchiero2020deep,cheng2025interpolation}, however these results come at a cost as they require the user to have active control of the vector field of the Neural ODE (i.e.\ non-autonomous or controlled Neural ODEs) which is effectively the case where $I_d=\infty$.

One might already know that autonomous-flow-based generation can be represented as a single \textit{non-autonomous flow}, i.e., where the vector field in~\eqref{eq:autonomous_ODE} is allowed to be time-dependent but undergoes only finitely many changes in direction, which directly allows the conversion of our theory of autonomous-flow-based generation to time-dependent generation. This means one can write any autonomous-flow-based generation as time-one of a time-dependent vector field, and it is common to use a time-dependent vector field in Neural ODEs~\cite{chen2019neuralordinarydifferentialequations}. Moreover, one can write any $C^1$-diffeomorphism isotopic to the identity as a time-one of a time-dependent vector field \cite{Banyaga_1997__DiffeoBook}, so time-dependent vector fields do not lack expressive power.
On the other hand, \textit{controlled Neural ODEs}~\cite{kidger2020neural,cuchiero2020deep,cirone2023neural}, which mimic controlled rough paths~\cite{morrill2021neural,pmlr-v235-walker24a} can potentially adapt their vector fields arbitrarily often over time, resulting in a higher parametric complexity but more expressive power. Moreover one can also write any autonomous flow using a one-dimensional control. Now one might wonder:
\begin{question}
\label{eq:Main_Question}
Is there any advantage of considering the autonomous-flow-based generation at all?
\end{question}
The class of functions being used matters; if one only wants to learn or approximate a group of homeomorphisms or diffeomorphisms for example, defining additional controls doesn't help and the machinery in rough analysis is then useless. 
On the other hand, despite time-dependent vector fields, the autonomous restriction, which produces building blocks, might be easier to consider in practice, for example in kinematic robotics, it is easier to think of autonomous vector fields for each screw motion, but time-dependent vector fields might be too complicated and unnecessary to consider. 
Moreover, autonomous blocks have much more algebraic structure so one can utilize the tools in Lie groups, diffeomorphism groups, exponential maps, fragmentation, generation of groups, splitting methods, etc. 
Moreover, for non-autonomous flows, one doesn't have the nice one-parameter group identity, which yields a stationary transport or conservation equation~\cite{LeeEtAl2026}. 
In the case of approximating with Neural ODEs, the only difference is that the same parameter $\theta$ in a time-independent vector field $V_\theta$ determines the whole dynamics throughout the entire interval, but for a time-dependent vector field, the network must additionally model the dependence on $t$. One has to mention that in high dimensions, changing $d$ to $d+1$ is negligible and there is no advantage in considering autonomous vector fields. 

We show that autonomous-flow-based generation has enough expressivity and as a consequence we approximate them with Neural ODEs, as this is the natural Neural Network framework for this context.
\subsection{Main Contributions}
\subsubsection{Compactly Supported$_\id$:}
Our negative result (\cref{thrm:NonEmbedability}) shows that in autonomous-flow-based generation for Neural ODEs, a smaller compact support$_\id$ makes it nowhere dense in the space of the larger support$_\id$. Moreover, we prove that for the same support$_\id$, it is meagre (\cref{Main_Negative_Theorem}).

Our second result (in \cref{thrm:Uniform_Bound__Diff_Flows}) shows for any diffeomorphism $\varphi \in \Diff[\Rd][\zd][\infty] $, there exists a representation of it as a composition of at most $I_d \geq 1$  time-one autonomous flows such that $I_d$ only depends on the dimension $d$ and doesn't depend on $\varphi$.

Our result for diffeomorphisms (\cref{prop:Universality_of_Neural_ODEs__SmoothCase}) shows that by only considering composition, Neural ODEs of the same support$_\id$ are universal in the space of $\Diff[\Rd][\zd][\infty]$ quantitatively.
Our result for homeomorphisms (\cref{thrm:General_Universality}) shows that again by considering composition, autonomous Neural ODEs are universal in the space of $\Homeo[\Rd][\zd]$ for $d \geq 5$.

Finally, we prove that one can approximate compactly supported Lipschitz functions between arbitrary dimensions (\cref{cor:Optimal_approximation_general}).  
\subsubsection{Defined on the Cube:}
Our first result (\cref{thrm:ZOD_MainNegative_CompactManifold}) shows that the class of single autonomous neural ODEs is nowhere dense.

Our second result (\cref{prop:ZOD_Universality_of_Neural_ODEs__SmoothCase}) shows that by considering composition, Neural ODEs are universal in the space of diffeomorphisms on the cube.
 
\section{Preliminaries}
\label{s:Prelim}

\subsection{Background}
\label{s:Prelim__ss:Background}

\paragraph{Isotopy and Orientation-Preserving:} 
\label{s:Prelim__ss:Background___sss:Homeo} 
We will say that two homeomorphisms are \textbf{isotopic} if one can be transformed continuously into the other with homeomorphisms.
\begin{example}
A simple example of an isotopy on $\mathbb{R}^2$ is given by:
\[
H_t(x,y)
=
\begin{pmatrix}
\cos(t\theta) & -\sin(t\theta)\\
\sin(t\theta) & \cos(t\theta)
\end{pmatrix}
\begin{pmatrix}
x\\
y
\end{pmatrix},
\qquad t\in[0,1].
\]
For every $t\in[0,1]$, the map $H_t$ is a homeomorphism.
Thus, $(H_t)_{t\in[0,1]}$ is an isotopy from the identity map to a rotation.
\end{example}
A $C^k$-diffeomorphism is \textbf{orientation-preserving} if it preserves local orientations, equivalently, if its Jacobian determinant is positive in oriented coordinates;
for homeomorphisms it is formulated using local degree or local orientation classes, cf.~\cite{Hirsch1976}. In Euclidean space, it can be considered as being isotopic to the identity as the set of orientation-preserving homeomorphisms and the identity component are the same.
\begin{example}[Orientation-Preserving Homeomorphisms From Computer Vision (Rotations)]
\label{ex:ON}
\hfill\\
Standard multidimensional examples arising in rotation invariances in computer vision, e.g.~\cite{lui2012human,cohen2016group,thomas2018tensor}, include the linear transformation $\varphi:\mathbb{R}^d\ni x\mapsto Ox\in \mathbb{R}^d$ where $O$ is a $d\times d$ orthogonal matrix; in which case $\varphi$ is orientation-preserving if and only if $\det(O)=1$ and it fails to be precisely when $\det(O)=-1$.  

If $O$ is orientation-preserving; i.e.\ if $\det(O)=1$, then any such homeomorphism can be expressed as the solution to an ordinary differential equation (ODE) at time $1$; namely, $\varphi(x)= Ox=x_1^x$ where: 
\begin{equation}
\begin{aligned}
\label{eq:integral_curve_SpecialOrthonalGroup}
    \tfrac{d}{dt}\,x_t^x & = \mathfrak{o}x_t^x
\\
    x_0^x & = x,
\end{aligned}
\end{equation}
where $O=\exp(\mathfrak{o})$ for some $d\times d$-skew-symmetric matrix $\mathfrak{o}$; where $\exp$ is the matrix exponential.  
This can be seen by observing that the ODE solution to the ODE~\eqref{eq:integral_curve_SpecialOrthonalGroup} is given by the curve $x_{\cdot}^x=(x_t^x)_{t\ge 0}$ where 
\[
x_t^x = \exp(t\mathfrak{o})\,x
.
\]
This connection is the starting-point of the theory of finite-dimensional Lie groups; cf.~\cite{helgason1979differential}.  Indeed, finite-dimensional Lie groups induce the prototypical and simplest classes of ``well-behaved'' homeomorphisms on $\mathbb{R}^d$ in this way; Note that one can make any rotation compactly supported$_\id$ by multiplying its vector field by a suitable smooth bump function.
\end{example} 
\subsection{Notation}
\label{s:Prelim__ss:Notation}
\subsubsection{Used notations and definitions:}
For reference, we collect the notation and definitions used in this paper:
\begin{itemize}
    \item 
        Let $\mathbb{N}\eqdef \{0,1,2,\dots,\}$ , $ \mathbb{N}_+\eqdef \{n\in \mathbb{N}:\, n>0\}$ and for $N \in \mathbb{N}_+$ denote $[N] \eqdef \{1 , \dots , N\}$ , and let $\R_+$ be the set of positive real numbers.
    \item 
        For a Lipschitz function $f$ we denote its Lipschitz constant as $L^f$.
    \item 
        We denote the $i$-th component of a function $f:\Rd \rightarrow \R^{D}$ where $f = (f_1,\dots,f_D)$ as $(f)_i \eqdef f_i$ for $i \in [D]$.
    \item 
        Given $I$ functions $\{f_1,\dots,f_I\}$, the iterated composition operator $\bigcirc$ maps any finite set of composable functions $f_1,\dots,f_I$ to their composition $\bigcirc_{i=1}^I \,f_i\eqdef f_I\circ \dots \circ f_1$.
    \item 
        Given $E\subseteq \Rd$, $d$-tuple 
        \(
        \alpha = [\alpha_1, \alpha_2, \cdots, \alpha_d]^{\top} \in \mathbb{N}^d
        \) 
        and functions $f(x):\R^d \rightarrow \R^D$, $g:\R^d\rightarrow \R$  and $\operatorname{id}:\Rd \rightarrow \Rd$ let:
    \begin{itemize}
        \item 
            $\| x \|_{l^\infty} \eqdef \text{max}_{i \in [d]} |x_i|,$
        \item 
            \(
            \|g\|_{L^\infty(E)}
            \eqdef
            \text{ess sup}_{x \in E} \, |g(x)|,
            \)
        \item 
            \(
            \|f\|_{L^\infty(E; l^\infty)}
            \eqdef \|\text{ess sup}_{x \in E} \, |f_1(x)| \, , \dots, \, \text{ess sup}_{x \in E} \, |f_D(x)| \, \|_{l^\infty} 
            =
            \text{max}_{i \in [D]} \, \text{ess sup}_{x \in E} \, |f_i(x)| , 
            \)
        \item 
            $\|\cdot\|_2$ denotes the Euclidean norm,
        \item 
            \(
            \|\alpha\|_1 \eqdef |\alpha_1| + |\alpha_2| + \cdots + |\alpha_d|,
            \)
        \item 
            \( \partial^\alpha \eqdef \frac{\partial^{\alpha_1}}{\partial x_1^{\alpha_1}}
            \frac{\partial^{\alpha_2}}{\partial x_2^{\alpha_2}}
            \cdots
            \frac{\partial^{\alpha_d}}{\partial   x_d^{\alpha_d}},
            \)
        \item 
            \(
            \|g\|_{C^s(E)} = \max \left\{ \|\partial^\alpha g\|_{L^\infty(E)} 
            : \alpha \in \mathbb{N}^d \ \text{with}\ \|\alpha\|_1 \leq s \right\}.
            \)\footnote{Note that $\|g\|_{C^0}=\|g\|_{L^\infty}$}
        \end{itemize}
        
        \item 
            Let \(I=[0,\infty)\) or \(I=[0,T]\) for some \(T\geq 0\). A function:
            \[
            \omega\colon I\to[0,\infty),
            \]
            is called a \emph{modulus of regularity} if:
            \begin{enumerate}
                \item \(\omega(0)=0\);
                \item \(\omega\) is monotone increasing;
                \item \(\omega\) is concave.
            \end{enumerate}
\end{itemize}
Given a subset $P\subseteq M$ of a manifold $M$, for $k\in \mathbb{N}_+\cup \{\infty\}$:
\begin{itemize}
    \item 
        We denote the interior of $P$ by $int(P)$.
    \item 
        Given a function $f:M\to\R^D$, we denote its support by $\mathrm{supp}(f)\eqdef \overline{\{x\in M:\, f(x)\neq 0\}}$. For a function $f:M\to M$ its support$_\id$ is denoted as $\mathrm{supp}_\id(f)\eqdef \overline{\{x\in M:\, f(x)\neq x\}}$. 
    \item 
        Let $\Homeo[M][P], \Diff[M][P][k]$ be the set of homeomorphisms and $C^k$-diffeomorphisms in the identity component of $M$ while compactly supported$_\id$ on $P$. Let $\mathcal{H}(M,P)_{+}, \mathrm{Diff}^k(M,P)_+$ be the set of orientation-preserving homeomorphisms and $C^k$-diffeomorphisms of $M$ compactly supported$_\id$ on $P$. For simplicity, when there is no confusion, we might write $\Homeo[P][P]$ and $\Diff[P][P][k]$ as $\Homeo[P]$ and $\Diffk[P][k]$.
    \item 
       Let $\calX^k(M,P)$ and $\calX^0(M,P)$ be the set of $C^k$ and \textit{Lipschitz} continuous vector fields on $M$ compactly supported$_\id$ on $P$ respectively. For simplicity, when there is no confusion, we might write $\calX^k(P,P)$ and $\calX^0(P,P)$ as $\calX^k(P)$ and $\calX^0(P)$.
    \item 
        \label{Cauchy_Prob__Vect}
        For a given vector field $V \in \calX^0(M,P)$, let $\mathrm{Flow}(V)$ denote the time-one map of the corresponding $C^1$ integral curve $\gamma$ whose ODE system can be written in coordinates as:
        \begin{equation}
        \label{Cauchy_VectorField}
        \begin{split}
        \begin{cases}
            \dot{x}_t^x & = V(x_t^x),
        \\
            x_0^x & = x.
        \end{cases}
        \end{split}
        \end{equation}
    \item 
        Fix $I\in \mathbb{N}_+$, $k\in \mathbb{N}$. We say $\varphi\in \Homeo[M][P][I]$ if there exist vector fields $V_1,\dots,V_I\in \calX^0(M,P)$ such that:
            \begin{equation}
            \label{eq:defn:HomeoStructure__Representation_Homeo}
                    \varphi 
                = 
                    \bigcirc_{i=1}^I\,
                        \Flow(V_i)
            .
            \end{equation}
        and $I$ is minimal among all of the possible representations. We say $\varphi\in \Diff[M][P][k,I]$ if there exist vector fields $V_1,\dots,V_I\in \calX^k(M,P)$ such that:
            \begin{equation}
            \label{eq:defn:HomeoStructure__Representation}
                    \varphi 
                = 
                    \bigcirc_{i=1}^I\,
                        \Flow(V_i)
            .
            \end{equation}
        and $I$ is minimum among all of the possible representations.
        For simplicity, when there is no confusion, we might write $\Homeo[P][P][I]$ and $\Diff[P][P][k,I]$ as $\HomeoI[P][I]$ and $\Diffk[P][k,I]$.
    \item 
        A function $f: M \rightarrow M$ is called \textbf{flowable} if there exists a vector field $V$ on $M$ such that $f=\Flow(V)$.
\end{itemize}
Let $(M,g)$ be a connected Riemannian manifold:
\begin{itemize}
    \item 
        The length of a piecewise $C^1$ curve $\gamma\colon[a,b]\to M$ is defined by:
        \[
        L_g(\gamma)
        \eqdef
        \int_a^b
        \sqrt{
        g_{\gamma(t)}
        \bigl(\dot{\gamma}(t),\dot{\gamma}(t)\bigr)
        }
        \,dt.
        \]
    \item
        The distance induced by the Riemannian metric $g$ is:
        \[
        d_g(p,q)
        \eqdef
        \inf_{\substack{
        \gamma\colon[a,b]\to M\\
        \gamma(a)=p,\ \gamma(b)=q
        }}
        L_g(\gamma),
        \qquad p,q\in M,
        \]
        where the infimum is taken over all piecewise $C^1$ curves joining
        $p$ to $q$. For the Euclidean metric on $M=\Rd$: $d_g = \| p - q \|_2$.
    \item 
        The norm induced by metric $g$ is:
        \[
        \|v\|_g \eqdef \sqrt{g_p(v,v)}, 
        \qquad
        v \in T_pM.
        \]
    \item 
            We denote the ball in $M$ with radius $r$ around point $p$ as $B_M(p,r) \eqdef \{x\in M |\ d_g(x,p)<r\}$.
    \item 
        For two homeomorphisms $f,g$ on $M$ define:
        \[
        d_1(f,g) 
        \eqdef
        \sup_{x\in M}
        \Bigl\{
        d_g\bigl(f(x),g(x)\bigr),
        d_g\bigl(f^{-1}(x),g^{-1}(x)\bigr)
        \Bigr\}.
        \]
    \item 
        \[
        d_{C^0}(f,g)
        \eqdef
        \sup_{x\in M} d_g\bigl(f(x),g(x)\bigr).
        \]
\end{itemize}

\subsubsection{Neural ODEs}
Let $\Delta\in\N_+$ and consider a multi-index 
$\mathbf{d}\ \eqdef\ [d_{1},\ldots,d_{\Delta+1}]^{\top}$ such that $d_i \in \N_+$ for $i \in [\Delta+1]$.
The class $\mathcal{NN}(\mathbf{d})$ consists of all multilayer perceptrons with a Lipschitz activation function $\sigma \in C(\R)$ ($\sigma$-MLPs):
\[
\Phi:\ \mathbb{R}^{d_{1}}\to\mathbb{R}^{d_{\Delta+1}},
\]
admitting the following iterative representation:
\begin{equation}\label{eq:relu-mlp}
\begin{aligned}
\Phi(\mathbf{x}) &= \mathbf{W}^{(\Delta)} \mathbf{x}^{(\Delta)} + \mathbf{b}^{(\Delta)},\\[2pt]
\mathbf{x}^{(l+1)}\ &\eqdef\ \sigma \bullet\!\bigl(\mathbf{W}^{(l)}\mathbf{x}^{(l)}+\mathbf{b}^{(l)}\bigr),
&\qquad &\text{for } l=1,\ldots,\Delta-1,\\
\mathbf{x}^{(1)}\ &\eqdef\ \mathbf{x}\, .
\end{aligned}
\end{equation}
Here, for $l \in [\Delta]$, $\mathbf{W}^{(l)}$ is a $d_{l+1}\times d_{l}$ matrix and 
$\mathbf{b}^{(l)}\in\mathbb{R}^{d_{l+1}}$, and $\sigma\ \bullet$ denotes
componentwise application of the $\sigma$ function.

A $\sigma$-neural ODE of respective depth and width $\Delta,W\in \mathbb{N}^+$ is a flowable homeomorphism $\Psi \in \Homeo[\Rd]$ for which there exists a $\sigma$-MLP $\Phi:\mathbb{R}^d\to \mathbb{R}^d$ of depth $\Delta$ and width $W$ such that $\Psi = \Flow(\Phi)$. 
        We denote the class of all $\sigma$-Neural ODEs compactly supported$_\id$ on $P \subseteq \Rd$ as $\mathrm{NODE}_\sigma(\Rd,P)$. For simplicity, when there is no confusion, we might write $\mathrm{NODE}_\sigma(P,P)$ as $\mathrm{NODE}_\sigma(P)$.
\section{First Case : Compactly Supported$_\id$ on $\zd$}
\label{s:Main}

\subsection{Negative Results}
\label{s:Main__ss:Negative}
In this subsection, we show that the set $\NODE_\sigma(\Rd,\zd)$ is meagre in $\Homeo[\Rd][\zd]$.
The reason that the following theorem is stated in $\Rd$ and not on a general $C^\infty$ manifold is that $\Homeo[\Rd][\Rd] = \mathcal{H}(\Rd,\Rd)_+$ and~\cref{Gen_CSupp} yields an orientation-preserving homeomorphism, not necessarily in the identity component. Note that naturally, time-one flows are in the identity component. The theorem is as follows (cf.~\cite[Theorem A]{BonomoVarandas2020}):

\begin{theorem}
\label[theorem]{thrm:NonEmbedability}
Let $d\in \mathbb{N}_+$ with $d>1$. Fix compact sets $K \subseteq Q \subseteq \Rd$ with path connected interiors such that $K$ is compactly contained in $\inter(Q)$ and $K$ has non-empty interior. 
Then the subset of flowable $\Homeo(\Rd,K)$ is nowhere dense in $\Homeo[\Rd][Q]$ in $C^0$-topology.
\end{theorem}
\begin{proof}:
Choose an arbitrary flowable $f \in \Homeo[\Rd][K]$.
By \cref{ex_rec} (for $M=K$) there exists $x\in R(f)$.

First, consider the case $R(f) \backslash \text{Fix}(f) \neq \emptyset$ and take $x\in R(f) \backslash \text{Fix}(f)$. As calculation in $d_1$ distance results in $d_{C^0}$ distance, by~\cref{Gen_CSupp},  
there exists an $\frac\varepsilon2$-$C^0$-perturbation $f_1 = f \circ h$ such that $f_1^k(x)=x$ for some $k>1$ and $f_1 \in \Homeo[\Rd][\tilde K]$ where $K \subset \tilde K$ is compact. 
Choose $r$ small enough such that $\mathcal{B}=\{B(f_1^i(x),r):1\leq i \leq k \}$ forms a pairwise disjoint collection of balls inside $Q$.

Set $p_i \eqdef f_1^i(x)$. Choose small balls $U_i$ and compact closed balls $V_i$ such that $p_i \in V_i \subset U_i \subset \overline{B(p_i,r)}$ and $f_1(V_i) \subset B(p_{i+1},r)$; and corresponding orientation-preserving coordinate charts $\psi_i: U_i \rightarrow \R^d$ such that $\psi(p_i)=0$. Define $g_i: U_i \rightarrow U_{i+1}$, $g_i \eqdef \psi_{i+1}^{-1}(\lambda \psi_i(y))$ for $0<\lambda<1$ on $U_i$, thus $g_i(p_i) =p_{i+1}$ and 
in coordinates $\|\psi_{i+1}(g_i(y))-\psi_{i+1}(g_i(z))\|_2 = \lambda \|\psi_i(y) - \psi_i(z)\|_2$ so $g_i$ is smooth and contracting. Set $c_i : f_1(U_i) \rightarrow U_{i+1}$, $c_i \eqdef g_i \circ f_1^{-1}$ which is a homeomorphism from $f_1(V_i)$ to the range of $g_i$; 
Define $\tilde c_i : f_1(V_i) \rightarrow c_i(f_1(V_i))$, $\tilde c_i \eqdef c_i|_{f_1(V_i)}$. The closed sets $f_1(V_i)$ and  $c_i(f_1(V_i))$ are flat topological closed balls. This is because one can consider a homeomorphism $\varphi: B(0,2) \rightarrow U_i$ such that $\varphi|_{\overline {B(0,1)} }$ is a homeomorphism from $\overline {B(0,1)}$ to $V_i$. Then by the way the functions $\varphi, f_1$ and $c_i$ are constructed, the homeomorphism $\eta \eqdef  c_i  \circ f_1 \circ \varphi|_{\overline{B(0,1)}}$:
\[
\overline{B(0,1)} \xrightarrow{\varphi|_{\overline{B(0,1)}}} V_i \xrightarrow{f_1} f_1(V_i) \xrightarrow{c_i} c_i(f_1(V_{i})),
\]
can be extended to a homeomorphism from the neighborhood $B(0,2)$ of $\overline{B(0,1)}$:
\[
B(0,2) \xrightarrow{\varphi} U_i \xrightarrow{f_1} f_1(U_i) \xrightarrow{c_i} U_{i+1}.
\]
This proves that $c_i(f_1(V_{i}))$ is a flat topological ball. Similarly, $f_1(V_{i})$ is a flat topological ball (without composing $c_i$, i.e. $\tilde \eta \eqdef  f_1 \circ \varphi|_{\overline{B(0,1)}}$). 

Then by \cite[Theorem 1.9.]{GOLDSTEIN2025125681} one can obtain homeomorphisms $H_i$ that are the identity outside of $B(p_{i+1},r)$ and are equal to $\tilde c_i$ on $f_1(V_i)$. Now note that $H_i \circ f_1 = g_i$ on $V_i$ and $H_i \circ f_1 = f_1$ outside of $B(p_{i+1},r)$. As the balls in $\mathcal{B}$ are disjoint, $f_2 \eqdef \bigcirc_{i=1}^k (H_i) \circ f_1$  is a homeomorphism which is equal to $f_1$ outside of $\mathcal{B}$, $f_2^i(x)$ is a periodic attractor for $f_2^k$ and $f_2^k(p_i)$ are isolated fixed points of $f_2^k$ for each $i \in \{1,\dots,k\}$. Having isolated attracting points yields a homeomorphism which is not flowable and by choosing $r$ sufficiently small, one gets $d_{C^0}(f_2,f_1)=\|f_2-f_1\|_{C^0}<\frac\varepsilon2$. 

This shows that such homeomorphisms $\{f_2\}$ form a dense set of homeomorphisms compactly supported$_\id$ on $Q$ which are not flowable and they have (non-fixed) periodic points of period $k>1$.

Note that for each homeomorphism $f_2$, there exists a sufficiently small neighborhood $U$ of $x$ such that ${f_2^k}_{|U}$ has no fixed point on $\partial U$.
For every $C^0$-Perturbation $g$ of $f_2$ in $\Homeo[\Rd][Q]$, one has that 
$0 \neq \text{deg}(f_2^k(x)-x, U ,0) =
\text{deg}(g^k(x)-x, U ,0)$, which means, there exists a point $z$ such that $g^k(z)-z=0$, or in other words, $g^k$ has a fixed point in $U$ and no fixed point on the boundary of $U$. Thus,
$g$ has a periodic point of period $k > 1$ that is not contained in a continuum subset of the set of $k$-periodic points.
It means $f_2$ has a $C^0$-open neighbourhood.
This proves the openness.

Now consider $x\in R(f) \subseteq \text{Fix}(f)$. 
Take a point $x \in \mathrm{int}(Q \backslash K)$ and $0 < r < \varepsilon$ small enough such that $B(x,r) \subseteq \mathrm{int}(Q \backslash K)$, then consider $\phi$ the time-one flow of the vector field $\rho \cdot R$, where $R$ is the vector field whose time-one map gives the $\pi$ rotation around the point $x$ and $\rho$ is a smooth bump function equal to 1 on $B(x,\frac{r}{2})$ and zero outside $B(x,r)$. The composition $f_1 \eqdef f \circ \phi \in \Homeo[\Rd][Q]$ satisfies $d_{C^0}(f_1,f)=\|f_1-f\|_{C^0}<\varepsilon$  having a period-two point (because $f$ is the identity outside $K$), which is the condition of having a non-fixed periodic point. This finishes the proof.
\end{proof}
\begin{theorem}
\label[theorem]{Main_Negative_Theorem}
    Let $d\in \mathbb{N}_+$ with $d>1$, then the subset of flowable $\Homeo[\Rd][\zd]$ is meagre in $\Homeo[\Rd][\zd]$ in $C^0$-topology.
\end{theorem}
\begin{proof}:
    Let $K_N \eqdef [\frac{1}{N}, 1 - \frac{1}{N}]^d$. By definition for $N \geq 3$:
    \[
    \Homeo[\Rd][\zd] = \bigcup_{K_N} \Homeo[\Rd][K_N],
    \]
    and by \cref{thrm:NonEmbedability}, for any compact set $K_N \subseteq K_{N+1}$, each flowable $\Homeo[\Rd][K_N]$ is nowhere dense in $\Homeo[\Rd][K_{N+1}] \subseteq \Homeo[\Rd][\zd]$, so the set of flowable $\Homeo(\Rd,\zd)$ is meagre in $\Homeo[\Rd][\zd]$.
\end{proof}
This means for any $d\in \mathbb{N}_+$ with $d>1$ and Lipschitz activation function $\sigma$, $\NODE_\sigma(\Rd,\zd)$ which is a subset of flowable $\Homeo(\Rd,\zd)$ is meagre in $\Homeo[\Rd][\zd]$.
Note that meagreness does not rule out universal approximation.
Note that by the above theorem, the subset of flowable $\Homeo[\Rd][\zd]$ is meagre in itself and as a result, it is not a Baire space. On the other hand, as the complete metric space $\Homeo(\Rd,\zod)$ is a Baire space~\cite[Corollary 25.4]{willard2004general}, the complement of flowable $\Homeo(\Rd,\zd)$ is a dense subset~\cite[Theorem 9.1.]{Oxtoby1971Cat}, and thus a potential for universal approximation. 
\subsection{Positive Results:}
\label{s:Main__ss:Positive}
\subsubsection{Quantitative Result:}
\label{s:Main__ss:Positive___sss:Quantitative}
Using the following theorem, we give an order of approximation for any $\Diff[\Rd][\zd][\infty]$ with autonomous Neural ODEs having the same support$_\id$ but using composition in \cref{prop:Universality_of_Neural_ODEs__SmoothCase}.
\begin{theorem}[Finite Composition of Flows for Diffeomorphisms]
\label[theorem]{thrm:Uniform_Bound__Diff_Flows}
    One can represent any diffeomorphism $\varphi \in \Diff[\Rd][\zd][\infty]$ as a composition of at most $I_d$ autonomous flows where $I_d \in \N_+$ only depends on the dimension $d$.
\end{theorem}
\begin{proof}:
By~\cref{lem:decomposition}, $\varphi \in \Diff[\Rd][\zd][\infty]$ has a representation. By~\cref{thrm:structure} we conclude that among all the representations, there is one with $I$ autonomous flows where $I \le I_d$. This means one can write any diffeomorphism as a composition of at most $I_d$ flows $\varphi_i \in F(\Rd,\zd)$.
\end{proof}

\begin{theorem}[Universal Approximation by Deep Neural ODEs]
\label[theorem]{prop:Universality_of_Neural_ODEs__SmoothCase}
Let $\varphi \in \Diff[\Rd][\zd][\infty]$ and $N,L, k\in \N_+$ with $L$ fixed;
then,
there exist ReLU Neural ODEs $\{\Psi_i=\mathrm{Flow}(\Phi_i)\}_{i=1}^{I}$ such that $ \Psi \eqdef \bigcirc_{i=1}^{I}\,\Psi_i$ satisfies the approximation guarantee:
\begin{equation}
\label{eq:Universality_of_Neural_ODEs__2__}
\begin{split}
\big\| \varphi - \Psi \big\|_{L^\infty(\R^d; l^\infty)} \in
    O\big( (NL)^{-2k/d} \big).
\end{split}
\end{equation}
Moreover, $\Psi \in \Homeo[\Rd][\zd]$, with Lipschitz constant at most $\prod_{i=1}^{I} e^{L^{\Phi_i}}$ and $\Phi_1,\dots,\Phi_I$ are ReLU MLPs of width at most $d + 2d^2 + 17 k^{d+1}3^dd^2(N+2) \log_2(8N)$ and depth at most $18k^2 (L+2) \log_2(4L) + 2d+ 3 $. Note that $L^\Psi$ depends on the parameter $N$. In terms of number of parameters $P$ when $L$ is not fixed, one can write:
\begin{equation}
\label{Smooth_InParam}
    \| \Psi-\varphi\|_{L^\infty( \Rd ; l^\infty )} 
    \in
        \mathcal{O} \big( (\frac{P}{\log P})^{-2k/d} \big).
\end{equation}
\end{theorem}
\begin{proof}:
As $\varphi \in \Diff[\Rd][\zd][\infty]$ by~\cref{thrm:Uniform_Bound__Diff_Flows} there exists $I \le I_d$ such that $\varphi \in \Diff[\Rd][\zd][k,I]$ for any $k \in N_+$. 
Then by~\cref{prop:Universality_of_Neural_ODEs__DifableCase} there exists $\Psi$ such that:
\begin{equation}
\label{eq:Universality_of_Neural_ODEs__2}
    \begin{split}
        \big\| \varphi - \Psi \big\|_{L^\infty(\R^d; l^\infty)} &\le
            \sum_{i=1}^{I} \left( \|\upsilon_i(n,L)\|_{l^\infty} \prod_{j=i}^{I} e^{L^{V_j}}  \right) \\
            & \le 
            85 (k+1)^d 8^k\max_{i,j}\{\|(V_i)_j\|_{C^k([0,1]^d)}\}
            (NL)^{-2k/d}
            e^{I_d\max_i\{L^{V_i}\}}
            I_d \\
            & \in \mathcal{O} \big( (NL)^{-2k/d} \big),
    \end{split}
\end{equation}
as $(\upsilon_i)_j = 85 (k+1)^d 8^k \|(V_i)_j\|_{C^k([0,1]^d)}(NL)^{-2k/d}$. \eqref{Smooth_InParam} is a direct result of~\cref{prop:Universality_of_Neural_ODEs__DifableCase}. 
\end{proof}
\begin{remark}
    Note that one can also use \cref{thrm:Universality_of_Neural_ODEs_RH} instead of~\cref{prop:Universality_of_Neural_ODEs__DifableCase} in the above theorem, which gives a Lipschitz constant $L^\Psi$ independent of $N$ but with worse approximation rates $\mathcal{O}(P^{-1/d})$ as in~\cref{Lip_InPar}. See also \cref{Table_Complexity}. 
\end{remark}
\subsubsection{Qualitative Result:}
We show that for $d\ge 5$, one can approximate $\Homeo[\Rd][\zd]$ with autonomous Neural ODEs having the same support$_\id$:
\begin{theorem}[Universal Approximation of Orientation-Preserving Homeomorphisms]
\label[theorem]{thrm:General_Universality}
Let $d\in \mathbb{N}_+$ and $d\ge 5$.  There exists a constant $I_d\in \mathbb{N}_+$ such that for every $\varphi\in \Homeo[\Rd][\zd]$ and every $\varepsilon>0$ there exist an $I\le I_d$ and ReLU Neural ODEs $\{\Psi_i=\mathrm{Flow}(\Phi_i)\}_{i=1}^{I}$ such that the homeomorphism $ \Psi \eqdef \bigcirc_{i=1}^{I}\,\Psi_i$ is Lipschitz and compactly supported$_\id$ on $\zd$ and satisfies the approximation guarantee:
\begin{equation}
\label{eq:Universality_of_Neural_ODEs}
         \big\| \varphi - \Psi \big\|_{L^\infty(\Rd; l^\infty)}
    \le
       \varepsilon
.
\end{equation}
\end{theorem}
\begin{proof}:
By~\cref{lem:Munkres_Connell_Bing}, one has for every $\varepsilon>0$, there exists a diffeomorphism $\varphi_{\varepsilon} \in \Diff[\Rd][\zd][\infty]$ satisfying the uniform approximation guarantee:\footnote{Recall the norm in \cref{lem:Munkres_Connell_Bing} is the uniform norm}
\begin{equation}
\label{eq:uniform_guarantee}
        \|
            \varphi_{\varepsilon}- \varphi 
        \|_{L^\infty(\Rd; l^\infty)}
    \le 
        \tfrac{\varepsilon}{2}
.
\end{equation}
By applying~\cref{thrm:Uniform_Bound__Diff_Flows,thrm:Universality_of_Neural_ODEs_RH}, 
there exist ReLU Neural ODEs $\{\Psi_i=\mathrm{Flow}(\Phi_i)\}_{i=1}^{I} $ for $I \le I_d$ such that the homeomorphism $ \Psi \eqdef \bigcirc_{i=1}^{I}\,\Psi_i$ is Lipschitz and compactly supported$_\id$ on $\zd$ and satisfies the approximation guarantee: 
\begin{equation}
\label{eq:Universality_of_Neural_ODEs__aprx__M}
        \big\| \varphi_\varepsilon - \Psi \big\|_{L^\infty(\Rd; l^\infty)}
    \le
        \tfrac{\varepsilon}{2}
.
\end{equation}
Inequalities \eqref{eq:uniform_guarantee} and \eqref{eq:Universality_of_Neural_ODEs__aprx__M} yields the theorem. 
\end{proof}

\subsection{Universal Approximation of compactly supported Lipschitz functions between arbitrary dimensions}
\label{s:Ramifications__ss:continuousapprox}
At first glance, the homeomorphisms for which the domain and image have the same dimension may seem to suggest that our universal flow-based generative models are overly restrictive and thus cannot approximate continuous functions between Euclidean spaces, locally on compact subsets as standard multilayer perceptrons do; cf.~\cite{hornik1989multilayer,cybenko1989approximation,funahashi1989approximate}.  However, this is not the case. Using an idea similar to Klee’s trick, cf.~\cite{KleesTrick}, we encode the graph of an arbitrary continuous function into a homeomorphism in a higher dimension, in particular, by embedding the function into a \textit{single} simple time-$1$ autonomous flow. 

Before proceeding, we recall that in all of the theorems in this paper, the support can be any bounded open set diffeomorphic to $\zd$.

Fix $d,D\in \mathbb{N}_+$.  
Suppose the Lipschitz function $f:\mathbb{R}^d\to \mathbb{R}^D$ is compactly supported on $\zd$. $f$ induces a compactly supported Lipschitz vector field $V_f:\mathbb{R}^{d+D}\to \mathbb{R}^{d+D}$, defined for each $(x,y)\in \mathbb{R}^{d+D}$ by:
\begin{equation}
\label{eq:SpecialAutonomousODE}
        V_f(x,y)
    \eqdef 
        \big(
            0
            ,
            f(x)\rho(y)
        \big)
,
\end{equation}
where $\rho$ is a smooth bump function equal to $1$ on $\overline{B_{\R^D}(0,\max_{x
\in \Rd}\|f(x)\|_2 + 1)}$ and zero outside the ball $B_{\R^D}(0, \max_{x
\in \Rd} \|f(x)\|_2 + 2)$. The solution $\Phi:[0,\infty)\times \mathbb{R}^{d+D}\to \mathbb{R}^{d+D}$ to the \textit{autonomous} ODE induced by $V_f$ is defined by:
\[
\frac{dx_t^{(x,y)}}{dt}=0,
\qquad
\frac{dy_t^{(x,y)}}{dt}=f(x_t^{(x,y)})\rho(y_t^{(x,y)}),
\mbox{ where } (x_0,y_0)=(x,y),
\]
can be solved as:
\[
\frac{dx_t^{(x,y)}}{dt}=0 \Rightarrow x_t^{(x,y)} = x \Rightarrow
y_t^{(x,y)} = y_0 + f(x) \int_0^t \rho(y_t^{(x,y)}) dt .
\]
This means that: 
\begin{equation}
    \Flow(V_f)(x,0)=(x,f(x)).
\end{equation}
and $\Flow(V_f)$ is compactly supported$_\id$.
Now, let $\pi_D^{d+D}:\mathbb{R}^{d+D}\to \mathbb{R}^D$ denote the canonical (linear) projection sending any $(x,y)\in \mathbb{R}^{d+D}$ to $y\in \mathbb{R}^D$ and let $\iota_{d}^{d+D}:\mathbb{R}^d\to \mathbb{R}^{d+D}$ denote the (linear) embedding sending any $x\in \mathbb{R}^d$ to $(x,0)\in \mathbb{R}^{d+D}$; 
Putting it all together we have that for all $x\in \Rd$:
\begin{equation}
\label{eq:map_to_flow__prototype}
        \pi_D^{d+D}
    \circ
        \Flow(V_f)
    \circ 
        \iota_d^{d+D}
        (x)
    =
        f(x)
.
\end{equation}
Applying~\cref{thrm:General_Universality} to $\Flow(V_f)$ for $d+D \ge 5$, we conclude that by post- and pre-composing, one can uniformly approximate any Lipschitz function compactly supported on $\zd$.  

However, for diffeomorphisms, we note that, the model in~\eqref{eq:map_to_flow__prototype} does not scale well when $D$ is large, since one would obtain an approximation rate of $\mathcal{O}(P^{-1/(d+D)})$, (where $P$ is the number of parameters cf.~\cref{Lip_InPar}),  which is significantly slower than the ``unconstrained'' optimal rate achievable by ReLU MLPs; cf.~\cite{yarotsky2018optimal}.  Instead, a rate nearly equal to the ``unconstrained'' ReLU MLP rate is possible if we alternatively approximate each component of the target function $f$ independently and then concatenate the resulting lifted flow-based approximations. Denoting $f=(f_1,\dots,f_D)$, we replace~\eqref{eq:map_to_flow__prototype} with the model:
\begin{equation}
\label{eq:map_to_flow}
        \bigoplus_{i=1}^D
        \,
        \Big(
                \pi_1^{d+1}
            \circ
                \Flow(V_{f_i})
            \circ 
                \iota_d^{d+1}
                (x)
        \Big)
    =
        \big(
            f_1(x),\dots,f_D(x)
        \big)
    =
        f(x)
.
\end{equation}
The advantage of the representation in~\eqref{eq:map_to_flow}, with its greater width, over the more ``naive narrow'' version in~\eqref{eq:map_to_flow__prototype}, is that each $\Flow(V_{f_i})$ performs its approximation in only one extra dimension beyond the physical dimension $d$. This design minimizes the approximation-theoretic difficulties that typically arise from high dimensionality.
Importantly, doing so achieves the minimax optimal approximation rates (cf.~\cite[Remark (ii), p.111]{SHEN2022101} and~\cite{yarotsky2017error}) for the \textit{lifted, and thus higher-dimensional} space $\R^{d+1}$ and nearly achieves the optimal rate on the original low-dimensional domain $\Rd$, at the cost of one extra ambient dimension, which may be inevitable due to the extra invertibility structure of the autonomous models studied herein.

\begin{corollary}[Approximation of compactly supported Lipschitz Functions by Linear Lifting]
\label[corollary]{cor:Optimal_approximation_general}
Let $d,D\in \mathbb{N}_+$, and $f:\mathbb{R}^d\to \mathbb{R}^D$ be Lipschitz and compactly supported on $\zd$.  For every $n\in \mathbb{N}_+$ and each $i\in [D]$ there exist ReLU Neural ODEs $\Psi_i:\mathbb{R}^{d+1}\to \mathbb{R}^{d+1}$ such that the \textbf{Latent} Neural ODE defined as:
\begin{equation}
\label{Latent_NODE}
    \Psi \eqdef \bigoplus_{i=1}^D(\pi_1^{d+1} \circ \Psi_i \circ \iota_d^{d+1}),
\end{equation}
satisfies
\[
        \big\|
            f
            -
                \Psi
        \big\|_{L^\infty(\Rd; l^\infty)}
    \in
        \mathcal{O}\big(
        \tfrac{d+1}{n}
        \big)
.
\]
Moreover, $\Psi$ is Lipschitz with a Lipschitz constant independent of $n$ and the ReLU MLPs parameterizing the vector fields defining each $\Psi_i$ have width and parameters at most $\mathcal{O}(D(d+1)^2(n+1)^{d+1})$ and depth at most $\mathcal{O}( \lceil \log_2(d+1) \rceil )$ for each $i\in [D]$.
\end{corollary}
\begin{proof}:
Recall in \eqref{eq:map_to_flow}, $V_{f_i}$ is compactly supported for each $i \in [D]$, by~\cref{thrm:Universality_of_Neural_ODEs_RH}
there exist Lipschitz ReLU Neural ODEs $\{\Psi_i = \Flow(\Phi_i)\}_{i=1}^D$ such that for $\varepsilon \in O\big(\tfrac{d+1}{n}\big)$ and each $i \in [D]$ one has:
\[
\|\Flow(V_{f_i}) - \Psi_i\|_{L^\infty(\R^{d+1}; l^\infty)} < \epsilon.
\]
By restricting the domain and only considering the last coordinate of the output, one has:
\[
\| \pi_1^{d+1} \circ \Flow(V_{f_i}) \circ \iota_d^{d+1}
    - \pi_1^{d+1} \circ \Psi_i \circ \iota_d^{d+1}\|_{L^\infty(\Rd)} < \epsilon.
\]
Recall from \eqref{eq:map_to_flow} that $f_i = \pi_1^{d+1} \circ \Flow(V_{f_i}) \circ \iota_d^{d+1}$.
By enlarging the depth and width of each $\Phi_i$ to be equal to $\max_{i \in [D]} \{\text{depth}(\Phi_i) \}$ and $\max_{i \in [D]} \{ \text{width}(\Phi_i) \}$ if one needs to have the same width and depth, one can use concatenation~\cite[Proposition 2.3.]{petersen2024mathematical}:
\[
\Psi \eqdef \bigoplus_{i=1}^D(\pi_1^{d+1} \circ \Psi_i \circ \iota_d^{d+1}),
\]
where $\Psi_i$ has width and parameters at most $\mathcal{O}(D(d+1)^2(n+1)^{d+1})$ and depth at most $\mathcal{O}( \lceil \log_2(d+1) \rceil )$ for each $i \in [D]$. This yields:
\[
\|f - \Psi\|_{L^\infty(\Rd; l^\infty)} < \epsilon
.
\]
Note that by lifting, projecting and concatenation, the function remains Lipschitz.
Finally, recall by \cref{thrm:Universality_of_Neural_ODEs_RH} the Lipschitz constant is independent of $n$.
\end{proof}
\begin{remark}
\label[remark]{Lip_InPar}
    To write the order in terms of parameters, as the number of parameters $P \in \mathcal{O} \left( D (d+1)^2 (n+1)^{d+1} \right)$ then one can write the upper bound in $O\big((d+1)(\tfrac{D(d+1)}{P})^{1/(d+1)}\big)$ which means for fixed $d,D$:
    \begin{equation}
        \big\|
            f
            -
                \Psi
        \big\|_{L^\infty(\Rd; l^\infty)}
    \in
        \mathcal{O}\big(
        P^{-1/(d+1)}
        \big).
    \end{equation}
\end{remark}

\begin{remark}[Optimality of the $\mathcal{O}(P^{-1/(d+1)})$ rate]
The question of the optimality of the rate $\mathcal{O}(P^{-1/(d+1)})$ is not known, as there are no available approximation-theoretic lower bounds, nor tools for establishing lower bounds, in our non-vector space setting.  However, it is likely that a rate of $\mathcal{O}(P^{-1/d})$ cannot be achieved while requiring that the core of the model is pre- and post- composition of a homeomorphism by linear maps, due to the need to lift in order to approximate general continuous functions.  
\end{remark}

Thus, Corollary~\ref{cor:Optimal_approximation_general} provides a quantitative and more explicit version of the very recent qualitative result of~\cite{de2025approximation}.

\section{Second Case : Defined on $\zod$}
\label{s:ZOD_Main}

\subsection{Negative Result:}
\label{s:ZOD_Main__ss:Negative}
We show that considering only one single Neural ODE is not universal and in particular nowhere in this case.
\begin{theorem}
\label{thrm:ZOD_MainNegative_CompactManifold}
    $\NODE_\sigma(\zod)$ is nowhere dense in $\Homeo[\zod]$ for $d \geq 2$.
\end{theorem}
\begin{proof}
    By definition, $\NODE_\sigma(\zod)$ is a subset of flowable $\Homeo[\zod]$.
    From~\cite[Theorem A]{BonomoVarandas2020} flowable $\Homeo[\zod]$ is nowhere dense in $\Homeo[\zod]$. This proves the theorem.
\end{proof}
This means only considering a single autonomous flow does not result in universality and one has to consider composition. In the following, we show diffeomorphisms are not dense in the space of homeomorphisms, which is an obstruction for any method using this technique. 
\begin{theorem}
\label[theorem]{DifnotdenseinH}
    The subset $\Diffk[\zod][1]$ is not dense in the space $\Homeo[\zod]$.
\end{theorem}
\begin{proof}:
Define $h \in \Homeo[[0,1]^2]$ such that it rotates $(0,1)^2$ around $(\frac{1}{2},\frac{1}{2})$ and shifts the boundary by $\frac{1}{2}$. In particular, let \(Q=[0,1]^2\). Parametrize its boundary counterclockwise by:
\[
\gamma\colon \mathbb{R}/4\mathbb{Z}\longrightarrow \partial Q,
\qquad
\gamma(s) \eqdef
\begin{cases}
(s,0), & 0\leq s\leq 1,\\
(1,s-1), & 1\leq s\leq 2,\\
(3-s,1), & 2\leq s\leq 3,\\
(0,4-s), & 3\leq s\leq 4.
\end{cases}
\]
Define a boundary homeomorphism \(b\colon\partial Q\to\partial Q\) by
\(
b(\gamma(s))
\eqdef
\gamma\left(s+\frac{1}{2}\pmod 4\right).
\)
In particular,
\(
b(0,0)=b(\gamma(0))=\gamma\left(\frac12\right)
=\left(\frac12,0\right).
\)
Set \(c \eqdef (1/2,1/2)\). Every \(x\in Q\setminus\{c\}\) admits a
unique representation:
\[
x=c+r(p-c),
\qquad
p\in\partial Q,\quad 0<r\leq 1,
\]
where
\(
r=2\lVert x-c\rVert_{\infty},
p=c+\frac{x-c}{r}.
\)
Define \(h\colon Q\to Q\) by:
\[
h(c)=c,
\qquad
h\bigl(c+r(p-c)\bigr)
=
c+r\bigl(b(p)-c\bigr).
\]
Then \(h\) is a homeomorphism of \(Q\), with inverse obtained by
replacing \(b\) with
\(
b^{-1}(\gamma(s))
=
\gamma\left(s-\frac12\pmod 4\right).
\)
Moreover,
\(
h(0,0)=\left(\frac12,0\right).
\)

    By~\cite[Lemma 4.2.]{Jakob2026}, a $C^1$-diffeomorphism on the cube maps faces to faces and vertices to vertices. Define $ f \in \Homeo[\zod]$, $f \eqdef h \times \mathrm{id}_{[0,1]^{d-2}}$. $f$ has uniform distance at least $\frac{1}{2}$ from any diffeomorphism $g \in \Diffk[\zod][1]$. In particular, 
\[
\lVert g-f\rVert_{L^\infty ( \zod ; l^\infty )}
\geq
\left\lVert g(0)-f(0)\right\rVert_{l^\infty}
\geq \frac12.
\]
Thus \(h\) cannot be uniformly approximated by diffeomorphisms of
the square.
\end{proof}
This shows one cannot use the method in the proof of~\cref{thrm:General_Universality} to show universality for homeomorphisms. 
\subsection{Positive Results:}
\label{s:ZOD_Main__ss:Positive}
We start this section by remarking that the class $\Diffk[\zod][\infty,I]$ for some $I>1$ is non-empty.  Thus the following theorems provide a non-vacuous statement for $I>1$.
\begin{remark}[Non-triviality]
\label[remark]{prop:ZOD-Non-trivial}
    There exists an $I\in \mathbb{N}_+$ with $I>1$ such that $\Diffk[\zod][\infty,I]$ is non-empty.
\end{remark}
\begin{proof}:
    The proof is similar to~\cref{prop:Non-trivial} but instead of working on $\Rd$, one should work on $\zod$.
\end{proof}

\label{s:ZOD_Main__ss:Positive___sss:Quantitative}
Using the following theorem, we show that compositions of autonomous Neural ODEs uniformly
approximate every smooth diffeomorphism of the cube that is smoothly
isotopic to the identity through diffeomorphisms of the cube, with
error $\mathcal{O}(P^{-1/d})$ for each fixed target, where $P$ is the
total number of parameters.
\begin{theorem}[Finite Composition of Flows for Diffeomorphisms]
\label[theorem]{thrm:ZOD_Uniform_Bound__Diff_Flows}
    One can represent any diffeomorphism $\varphi \in \Diffk[\zod][\infty]$ as a composition of autonomous flows.
\end{theorem}
\begin{proof}:
According to~\cite{glockner2025boundary} the cube is a simple polytope. By~\cite[Corollary 1.6.]{glockner2025boundary} the identity group $\Diffk[\zod][\infty]$ can be written as a composition of time-one autonomous flows.
\end{proof}
\begin{theorem}[Universal Approximation by Deep Neural ODEs]
\label[theorem]{prop:ZOD_Universality_of_Neural_ODEs__SmoothCase}
Let $\varphi \in \Diffk[\zod][\infty]$ and $k\in \N_+$;
then,
there exist ReLU Neural ODEs $\{\Psi_i=\mathrm{Flow}(\Phi_i)\}_{i=1}^{I}$ such that $ \Psi \eqdef \bigcirc_{i=1}^{I}\,\Psi_i$ satisfies the approximation guarantee:
\begin{equation}
\label{eq:ZOD_Universality_of_Neural_ODEs__2__}
\begin{split}
\big\| \varphi - \Psi \big\|_{L^\infty( \zod; l^\infty)} \in
    O\big( \frac{d}{n} \big).
\end{split}
\end{equation}
Moreover, $\Psi \in \Homeo[\zod]$, with Lipschitz constant at most $\prod_{i=1}^{I} e^{L^{\Phi_i}}$ and $\Phi_1,\dots,\Phi_i$ are ReLU MLPs of width at most $8d^2(n+1)^d$, depth at most $ \lceil log_2d \rceil + 4$ and at most $16d^2(n+1)^d$ free parameters. Note that $L^\Psi$ does not depend on the parameter $n$. In terms of number of parameters $P$, one can write:
\begin{equation}
\label{ZOD_Smooth_InParam}
    \| \Psi-\varphi\|_{L^\infty( \zod ; l^\infty )} 
    \in
        \mathcal{O} \big( P^{-1/d} \big).
\end{equation}
\end{theorem}
\begin{proof}:
As $\varphi \in \Diffk[\zod][\infty]$ by~\cref{thrm:ZOD_Uniform_Bound__Diff_Flows} there exists an $I$ such that $\varphi \in \Diffk[\zod][k,I]$ for any $k \in \N_+$. 
Then by~\cref{thrm:ZOD_Universality_of_Neural_ODEs_RH} there exists $\Psi$ such that:
\begin{equation}
\label{eq:ZOD_Universality_of_Neural_ODEs__2}
    \begin{split}
        \big\| \varphi - \Psi \big\|_{L^\infty(\zod; l^\infty)}
        &\le
         \sum_{i=1}^{I} 
            \left( \|\omega_i(\frac{d}{2n})\|_{l^\infty}
            \prod_{j=i}^{I} e^{L^{V_j}}  \right),\\
            & \le 
            \max_i \{ \|\omega_i(\frac{d}{2n})\| \}
            e^{I_d\max_i\{L^{V_i}\}}
            I_d \\
            & \in \mathcal{O} \big( \frac{d}{n} \big),
    \end{split}
\end{equation}
as $(\upsilon_i)_j = 85 (k+1)^d 8^k \|(V_i)_j\|_{C^k([0,1]^d)}(NL)^{-2k/d}$. as the number of parameters $P \in \mathcal{O} \left( d^2 (n+1)^{d} \right)$ then one can write the upper bound in $O\big(d(\tfrac{d}{P})^{1/d}\big)$ which means for fixed $d$:
    \begin{equation}
        \big\|
            \varphi
            -
                \Psi
        \big\|_{L^\infty(\Rd; l^\infty)}
    \in
        \mathcal{O}\big(
        P^{-1/d}
        \big).
    \end{equation}
\end{proof}

\section{Comparison of domains}
\paragraph{Negative Result:}
Considering only the cube results a stronger negative result (nowhere density) while considering compactly supported$_\id$ on $\zd$ only gives meagreness, which is not an obstruction for universality despite nowhere density.
\paragraph{Positive Result:}
Considering the compactly supported$_\id$ on $\zd$ case, one can not only approximate diffeomorphisms, but also homeomorphisms. This is possible as one can approximate homeomorphisms in high dimensions by diffeomorphisms. Unfortunately, on the cube, we proved diffeomorphisms are not dense in the space of homeomorphisms so we cannot apply this method. 
\paragraph{Corollaries:}
We could deduce universality of compactly supported$_\id$ homeomorphisms on $\zd$ as we could approximate homeomorphisms by diffeomorphisms in this setting. Unfortunately, arbitrary functions might not be tangent to the cube and one cannot use the same lifting method. As a result, one cannot use the same method to even prove universality for differentiable functions defined on the cube.

\acks{
\noindent The authors would like to thank~\href{https://sites.google.com/view/paulovarandas/}{Paulo Varandas} and \href{https://andrew-warren.github.io/}{Andrew Warren}.

A.\ Kratsios and H.\ Rouhvarzi would like to acknowledge financial support from NSERC Discovery Grant No.\ RGPIN-2023-04482 and No.\ DGECR-2023-00230.  They also acknowledge that resources used in preparing this research were provided, in part, by the Province of Ontario, the Government of Canada through CIFAR, and companies sponsoring the Vector Institute\footnote{\href{https://vectorinstitute.ai/partnerships/current-partners/}{https://vectorinstitute.ai/partnerships/current-partners/}}.
}

\appendix
\section{Negative Result Propositions and Lemmas for Compactly Supported$_\id$ Case}
\label{app:NegativeRes}
\begin{proposition}[Existence of a recurrent point]
\label[proposition]{ex_rec}
    For every compact set $M \subset \Rd$ and any $f\in \Homeo[\Rd][M]$ there exists a recurrent point in $M$.
\end{proposition}
\begin{proof}: Take $f\in \Homeo[M][M]$.
Since $M$ is compact, the set:
\[
\mathcal{A} = \{S \subseteq M : S\neq \emptyset, S \text{ closed}, f(S)=S \},
\]
is non-empty (e.g. $M \in \mathcal{A}$).  Order \(\mathcal A\) by inclusion.  Any totally ordered chain \(\{S_i\}\) in \(\mathcal A\) has:
\[
\bigcap_i S_i \;\neq\;\varnothing,
\]
by compactness, and \(\bigcap_i S_i\) is closed and \(f\)-invariant. By the dual form of Zorn’s Lemma there is a minimal element \(S_{\min}\in\mathcal A\), i.e.\ a nonempty closed invariant set containing no proper nonempty closed invariant subset.
Take any \(x\in S_{\min}\).  If \(x\) were not recurrent then there would exist a neighborhood \(x \in U\) and an integer \(N\) such that:
\[
f^n(x)\notin U,
\quad
\forall\,n\ge N.
\]
Set:
\[
\omega(x) \;=\; \bigcap_{N \geq 0} \overline{\{\,f^n(x):n\ge N\}}.
\]
Then \(\omega(x)\) is nonempty, compact, and \(f\)-invariant, but \(x\notin \omega(x)\), so \(\omega(x) \subsetneq S_{\min}\), contradicting minimality.  Thus \(x\) must be recurrent.  Since \(x\) was arbitrary in \(S_{\min}\), every point of \(S_{\min}\) is recurrent.
\end{proof}
Let \((M,g)\) be a Riemannian manifold of dimension at least \(2\), with distance \(d_g\) coming from a Riemannian metric and let $\phi$ be a flow on $M$ and:
    \[
    \mathcal{Z}(\phi) \eqdef \{x:\dot{\phi}(x)=0\}.
    \]
Given a non-zero vector \(Y\in T_xM\) , where
    \(x\notin\mathcal{Z}(\phi)\), we define the \emph{inclination} of
    \(Y\) (relative to \(\phi\)) to be the length of the normalized
    difference:
    \[
    \sigma(Y)
    \eqdef
    \left\|
        \frac{1}{\lVert Y\rVert_g}Y
        -
        \frac{1}{\lVert\dot{\phi}(x)\rVert_g}\dot{\phi}(x)
    \right\|_g.
    \]
If the angle between \(Y\) and \(\dot{\phi}(x)\) is \(\theta\),
\(0\leq\theta\leq\pi\), then:
\[
\sigma(Y)=2\sin\left(\frac{\theta}{2}\right).
\]
Note that \(0\leq\sigma\leq2\), that \(\sigma=0\) if and only if \(Y\)
points in the same direction as \(\dot{\phi}\), and that:
\[
\widetilde{Y}
=
\left(\frac{\lVert\dot{\phi}\rVert_g}{\lVert Y\rVert_g}\right)Y,
\]
is a vector parallel to \(Y\) with:
\[
\lVert\widetilde{Y}-\dot{\phi}\rVert_g
=
\sigma(Y)\lVert\dot{\phi}\rVert_g.
\]
We define the inclination of a curve \(\gamma\) to be the inclination of
its velocity vector \(\dot{\gamma}\).
\begin{lemma}
\label[lemma]{lem:Shub_lem9}
    Let \((M,g)\) be a smooth Riemannian manifold of dimension at least \(2\) with metric $g$ and distance \(d_g\) coming from a Riemannian metric. Given $\varepsilon > 0$ and a global flow $\phi$, suppose $\gamma$ is a $C^1$ curve in $M$ 
    (an embedded closed interval or circle) such that at each point $x$ in the image of $\gamma$
    one of the following conditions holds:
    \begin{enumerate}[(i)]
        \item $\|\dot{\phi}(x)\|_g < \varepsilon / 2$, or
        \item $x \notin \mathcal{Z}(\phi)$, and $\gamma$ has inclination $\sigma < \varepsilon / \|\dot{\phi}\|_g$ at $x$.
    \end{enumerate}
    Then, given any neighbourhood $U$ of the image of $\gamma$, there exists a flow $\psi$
    on $M$ satisfying:
    \begin{enumerate}[(a)]
        \item $\dot{\psi} = \dot{\phi}$ off $U$,
        \item $\|\dot{\psi} - \dot{\phi}\|_g < \varepsilon$ on $M$,
        \item $\gamma$ is (a segment of an) integral curve of $\psi$.
    \end{enumerate}
\end{lemma}
\begin{proof}:
    For the proof, see \cite[Lemma 9]{NiteckiShub1975}. Note that in the paper \cite{NiteckiShub1975} the manifold is assumed to be compact but the proof of this lemma also works on non-compact smooth manifolds, provided $\phi$ is a global flow. This is because the proof does not explicitly use compactness of the ambient manifold $M$ and it is a local construction near $\gamma$. This can also be deduced from the same paper as in \cite[Lemma 13]{NiteckiShub1975} the Riemannian manifold is just assumed to be of dimension at least 2.
\end{proof}
\begin{lemma}
\label[lemma]{lem:Shub_lem13}
    Let \((M,g)\) be an orientable smooth manifold of dimension at least \(2\) with Riemannian metric $g$ and distance \(d_g\) coming from the Riemannian metric. 
    Suppose a finite collection \(\{(p_i,q_i)\in M\times M:\, i=1,\ldots,k\}\) of pairs of points of \(M\) is specified and a compact subset $P \subset M$ with path connected interior is such that $p_i, q_i \in \inter(P) $ (for \(i=1,\ldots,k\) ), together with a small positive constant \(\delta>0\) such that:
    \begin{enumerate}[(a)]
        \item For each \(i\), \(d_g(p_i,q_i)<\delta\).
        \item If \(i\ne j\), then \(p_i\ne p_j\) and \(q_i\ne q_j\).
    \end{enumerate}
    Then there exists an orientation-preserving \(f\in \Diff[M][P][\infty]\) such that:
    \begin{enumerate}[(i)]
        \item \(d_g\bigl(f(x),x\bigr)<2\pi\delta\) for every \(x\in M\).
        \item \(f(p_i)=q_i\) for \(i=1,\ldots,k\).
        \item $f$ is compactly supported$_\id$ on $P$.
    \end{enumerate}
\end{lemma}
\begin{proof}:
We consider the ``suspension'' of the identity, by taking \(M\times S^1\) with the product orientation and
with the vector field:
\[
X(p,\theta)=2\pi\frac{\partial}{\partial\theta}.
\]
The induced flow is, of course:
\[
\phi_t(p,\theta)=(p,\theta+2\pi t).
\]
Its time-one map, \(\phi_1\), takes \(M\times\{0\}\) to itself and
induces the identity map there.

Given the points \(p_i,q_i\in M\), we consider the points:
\[
(p_i,\pi/2)\quad\text{and}\quad(q_i,3\pi/2),
\]
in \(M\times S^1\). We take, for each \(i\), a curve \(\gamma_i(t)\) in
\(M\), \(0\leq t\leq1\), of constant speed, joining \(p_i\) to \(q_i\),
of length \(<\delta\). Then the curve \(g_i\) in \(M\times S^1\) given by:
\[
g_i(t)=\left(\gamma_i(t),\pi t+\frac{\pi}{2}\right),
\qquad 0\leq t\leq1,
\]
has velocity vector:
\[
\dot{\gamma}_i+\frac{1}{2}X,
\]
whose inclination is \(<2\pi\delta\). We can change \(g_i\) slightly so
that
$\dot{g}_i(t)=\frac{1}{2}X$ at $t=0,1,$
and then use \cref{lem:Shub_lem9} to find a vector field \(Y\), with:
\[
\lVert Y-X\rVert_g<2\pi\delta,
\]
for which the curves \(g_i\) are segments of integral curves. But then
the flow of \(Y\) takes \((p_i,\pi/2)\) to \((q_i,3\pi/2)\), so that
\((p_i,0)\) and $(q_i,2\pi)=(q_i,0)$
are joined by an integral curve of \(Y\). Hence, the time-one map of the
flow of \(Y\) is a diffeomorphism of \(M\times\{0\}\), within
distance:
\[
\int_0^1\lVert Y-X\rVert_g\,dt<2\pi\delta,
\]
of the identity, and taking \(p_i\) to \(q_i\).

The above proof is reproduced verbatim from \cite[Lemma 13]{NiteckiShub1975}. We add that property (a) in \cref{lem:Shub_lem9} makes the vector field $Y - X$ compactly supported on a finite number of neighborhoods $\bigcup U_i$ of the curves $g_i$ joining $p_i$ to $q_i$, which also makes its time-one flow compactly supported$_\id$ on $M$; as $U_i$ were arbitrary neighborhoods around the curves, one can choose them sufficiently small that $\bigcup U_i \subset \inter(P)$; which makes the time-one flow of $Y$ to be compactly supported on $P$.  By the suspension construction, the ambient flow preserves the positive \(S^1\)-normal direction. Consequently, its induced return map \(h:M\to M\) is orientation-preserving.
\end{proof}
\begin{lemma}[$C^0$-closing lemma for compactly supported functions]
\label[lemma]{Gen_CSupp}
Let $f\colon M \to M$ be a compactly supported$_\id$ homeomorphism on the compact set $K \subset M$ with path connected interior, where $(M,g)$ is a $\mathcal{C}^\infty$ orientable manifold of dimension at least $2$ with Riemannian metric $g$, with distance $d_g$ coming from the Riemannian metric, and let $x_0\in M$ be a non-wandering point. Then for every $\varepsilon>0$ and compact set $P$ with path connected interior such that $K \Subset \inter(P)$ ($K$ is compactly contained in $\inter(P)$) there exists a compactly supported$_\id$ homeomorphism $g\colon M\to M$ on $P$ such that $x_0$ is a periodic point of $g$ and
$$d_1(f,g)<\varepsilon.$$
Moreover, if $f$ is orientation-preserving, then $g$ is also orientation-preserving.
\end{lemma}

\begin{proof}:
The Lemma is trivial when $x_0$ is a fixed point or a periodic point. Assume otherwise.\\
If $M$ is compact and $K=M$, then, by \cite[Theorem 4]{AnnaA_1996__C0closing}, the result follows. Suppose not. 

As $f$ is compactly supported$_\id$, $\mathcal{O}_f(x_0)$ is a subset of $K$. Note $f$ is uniformly continuous on $P$ and \cref{lem:Shub_lem9,lem:Shub_lem13} are true on any Riemannian manifold like $M$.

Let \(\varepsilon\) be a small positive constant. Since \(f\) is
uniformly continuous on $P$, there exists \(\eta\), with
\(0<\eta<\varepsilon\), such that
\[
d_g\bigl(f(x_1),f(x_2)\bigr)<\varepsilon
\qquad\text{if}\qquad
d_g(x_1,x_2)<\eta.
\]
Choose \(\delta\), with \(0<\delta<\eta/2\), such that
\[
d_g\bigl(f^{-1}(x_1),f^{-1}(x_2)\bigr)<\frac{\eta}{2}
\qquad\text{if}\qquad
d_g(x_1,x_2)<\delta.
\]
This is possible since \(f^{-1}\) is also uniformly continuous. \\
Let \(z_0\in B(x_0,\delta)\) be such that there exists
\(m\in\mathbb Z\), \(m\neq0\), satisfying
\[
f^m(z_0)=z_m\in B(x_0,\delta),
\qquad
z_0\neq x_0,
\qquad
z_m\neq x_0.
\]

Without loss of generality, we may assume that \(m>0\); otherwise, we
take \(z_m\) instead of \(z_0\). We also take the smallest \(m\)
satisfying the above conditions.

Suppose first that \(m\geq2\). Set
\[
z_i=f^i(z_0),\qquad i\in\mathbb Z,
\]
and consider the finite collection
\[
\bigl\{(p_i,q_i)\in M\times M:i=0,\ldots,m-1\bigr\},
\]
where
\[
\begin{aligned}
&p_0=x_0,\quad p_1=z_1,\quad p_2=z_2,\quad\ldots,\quad
p_{m-2}=z_{m-2},\quad
p_{m-1}=z_{m-1}=f^{-1}(z_m),\\
&q_0=z_0,\quad q_1=z_1,\quad q_2=z_2,\quad\ldots,\quad
q_{m-2}=z_{m-2},\quad
q_{m-1}=f^{-1}(x_0).
\end{aligned}
\]
as $m \geq 2$ all of these points are in $K \subset \inter(P)$ and the points \(p_i\) and \(q_i\) ($i=0,\dots,m-1$) satisfy:
\begin{enumerate}[label=\textup{(\alph*)}]
    \item \(p_i\neq p_j\) and \(q_i\neq q_j\) for
    \(0\leq i<j\leq m-1\);
    \item \(d_g(p_i,q_i)<\eta/2\) for \(i=0,\ldots,m-1\).
\end{enumerate}
According to~\cref{lem:Shub_lem13}, there exists an orientation-preserving diffeomorphism
\(h\colon M\to M\) with the following properties:
\begin{enumerate}[label=\textup{(\alph*)}]
    \item \(d_1(h,\operatorname{id})<\eta\);
    \item \(h(p_i)=q_i\) for \(i=0,\ldots,m-1\).
    \item $h$ is compactly supported$_\id$ on $P$.
\end{enumerate}
Set \(g=f\circ h\). Clearly, \(g\) is a compactly supported$_\id$ homeomorphism on $P$ and \(x_0\) is a
periodic point of \(g\). Moreover, since
\(d_1(h,\operatorname{id})<\eta<\varepsilon\),
\begin{align*}
d_1(f,g)
&=
\max_{x\in M}
\Bigl\{
d_g\bigl(f(x),(f\circ h)(x)\bigr),
d_g\bigl(f^{-1}(x),(h^{-1}\circ f^{-1})(x)\bigr)
\Bigr\}\\
&=
\max_{x\in M}
\Bigl\{
d_g\bigl(f(x),(f\circ h)(x)\bigr),
d_g\bigl(x,h^{-1}(x)\bigr)
\Bigr\}\\
&<\varepsilon.
\end{align*}
Suppose now that \(m=1\). Choose \(\varrho\), with
\(0<\varrho<\eta\), such that
\[
d_g\bigl(f^{-1}(x_1),f^{-1}(x_2)\bigr)<\frac{\eta}{2}
\qquad\text{if}\qquad
d_g(x_1,x_2)<\varrho.
\]
Choose \(\theta\) , with \(0<\theta<\varrho/2\) and small enough, such that
\begin{itemize}
    \item $d_g\bigl(f(x_1),f(x_2)\bigr)<\frac{\varrho}{2}
\qquad\text{if}\qquad
d_g(x_1,x_2)<2\theta,$
    \item $B(x_0,\theta) \subset \inter(P).$
\end{itemize}
Let \(y_0\in B(x_0,\theta)\) be such that there exists \(n>0\)
satisfying
\[
f^n(y_0)=y_n\in B(x_0,\theta),
\qquad
y_0\neq x_0,
\qquad
y_n\neq x_0.
\]
We take the smallest \(n\) satisfying these conditions. If \(n\geq2\),
the problem reduces to the previous case. If \(n=1\), consider the
points
\[
p_0=x_0,\qquad
p_1=y_1=f^{-1}(y_2),\qquad
q_0=y_0,\qquad
q_1=f^{-1}(x_0).
\]
The points $p_0,p_1,q_0,q_1$ satisfy:
\begin{enumerate}[label=\textup{(\alph*)}]
    \item \(p_0\neq p_1\) and \(q_0\neq q_1\);
    \item
    \begin{align*}
    d_g(p_0,q_0)
    &=d_g(x_0,y_0)
      <\theta
      <\frac{\varrho}{2}
      <\frac{\eta}{2},\\
    d_g(p_1,q_1)
    &=d_g\bigl(y_1,f^{-1}(x_0)\bigr)\\
    &=d_g\bigl(f^{-1}(y_2),f^{-1}(x_0)\bigr)
      <\frac{\eta}{2},
    \end{align*}
    since
    \begin{align*}
    d_g(x_0,y_2)
    &\leq d_g(x_0,y_1)+d_g(y_1,y_2)\\
    &=d_g(x_0,y_1)+d_g\bigl(f(y_0),f(y_1)\bigr)\\
    &<\theta+\frac{\varrho}{2}
      <\varrho.
    \end{align*}
    \item $p_0,p_1,q_0,q_1 \in \inter(P)$
\end{enumerate}

Again,~\cref{lem:Shub_lem13} gives an orientation-preserving diffeomorphism
\(h\colon M\to M\) with the following properties:
\begin{enumerate}[label=\textup{(\alph*)}]
    \item \(d_1(h,\operatorname{id})<\eta\);
    \item \(h(x_0)=y_0\) and \(h(y_1)=f^{-1}(x_0)\).
    \item $h$ is compactly supported$_\id$ on $P$.
\end{enumerate}
Set \(g=f\circ h\). Again, \(g\) is a compactly supported$_\id$ homeomorphism on $P$, \(x_0\) is a
periodic point of \(g\), and, as before,
\[
d_1(f,g)<\varepsilon.
\]

If $f$ is orientation-preserving then $g= f \circ h$ is also orientation-preserving. This is because $h$ is orientation-preserving from~\cref{lem:Shub_lem13}.
\end{proof}

\section{Positive Result Propositions and Lemmas for Compactly Supported$_\id$ Case}
\label{app:theorem}
\subsection{Qualitative Part}
\begin{lemma}[Munkres, Connell, Bing]
\label[lemma]{lem:Munkres_Connell_Bing}
Let \(d\geq 5\), and let \(\varphi \in \Homeo[\Rd][B(0,1)]\).
Then \(\varphi\) can be approximated uniformly by diffeomorphisms in $\Diff[\Rd][B(0,1)][\infty]$ .
\end{lemma}
\begin{proof}:
    Note that $\varphi|_{B(0,1)}$ is isotopic to the identity through an isotopy that fixes point-wise a neighborhood of the boundary of $B(0,1)$ by Alexander's Trick~(\cite{AlexandersTrick}).
    Apply~\cite[Lemma 2]{muller2014uniform} on $\varphi|_{B(0,1)}$. Because the diffeomorphisms and $\varphi|_{B(0,1)}$ are identity near the boundary of $B(0,1)$ one can extend them on $\Rd$ by taking them equal to the identity outside of $B(0,1)$. Every compactly supported$_\id$ diffeomorphism is orientation preserving as the sign of the determinant of the diffeomorphisms outside of the compact supported is $1$ and it cannot change sign. By~\cite[§6. Lemma 2.]{Milnor1997Topology} every orientation preserving diffeomorphism on $\Rd$ is isotopic to identity. This means the diffeomorphisms are in $\Diff[\Rd][B(0,1)][\infty]$.
\end{proof}
\subsection{Quantitative Formulations Part}
\subsubsection{Independent Lipschitz Constant for Neural ODE, but Worse Rates}
\label{s:Proof__ss:Step_2}
\begin{lemma}[Optimal Regular Approximation by Sample-Interpolating ReLU MLPs]
\label[lemma]{HongAnas}
Let \(f\) be a function from \([0,1]^d\) to \(\mathbb{R}\), and let
\[
\omega \colon [0,d] \to [0,\infty)
\]
be a modulus of regularity of \(f\). Then, for any \(n \in \mathbb{N}_{+}\),
there exists a ReLU MLP \(\Phi\) on \([0,1]^d\) with width at most
\(8d(n+1)^d\), depth at most \(\lceil \log_2 d \rceil+4\), and at most
\(16d(n+1)^d\) parameters, which satisfies the approximation guarantee
\[
\left\|f-\Phi\right\|_{L^\infty([0,1]^d)}
\leq
\omega\left(\frac{d}{2n}\right),
\]
as well as the sample-interpolation guarantee
\[
f(\mathbf{x})=\Phi(\mathbf{x})
\quad\text{for all}\quad
\mathbf{x}\in
\left\{
\frac{0}{n},\frac{1}{n},\ldots,\frac{n}{n}
\right\}^{d}.
\]
Furthermore, \(\omega\) is a modulus of regularity of \(\Phi\) on \([0,1]^d\).
\end{lemma}
\begin{proof}:
    For the proof, see~\cite[Theorem 4.1]{Hong_2024__ReLUMLPs}. 
\end{proof}
\begin{lemma}[Transfer: Universal Approximation of Vector Field to Flow]
\label[lemma]{lem:Vectorfield_to_Flow}
Let $n\in \N_+$ and $V = (V_1 , \dots , V_d) \in \calX^0(\Rd,\zd)$ with
modulus of regularity \(\omega_j:[0,d]\to\mathbb{R}\) for $j \in [d]$. Consider its time-one flow:
\begin{equation}
\label{Eq:timeonedlowof_V}
\begin{aligned}
        \varphi(x)
    & =
        x_1^x
\\
        x_t^x
    &=
            x
        +
            \int_0^t\,
                V(x_s^x)
            \,
            ds,
            \qquad
            0\le t\le 1,
\end{aligned}
\end{equation}
then there exists a ReLU MLP $\Phi: \R^d \to \R^d$ supported on $\zd$ with width at most $8d^2(n+1)^d + 2d^2 + d$, depth at most $\lceil \text{log}_2d \rceil + 7$, and at most $16d^2(n+1)^d + 6d^3 + 9d^2$ parameters such that the time-one flow (Neural ODE):
\begin{equation}
\label{eq_lem:decomposition__flow_representation___MLPVersion___RH}
\begin{aligned}
        \Psi(x)
    & =
        z_1^x,
\\
        z_t^x
    &=
            x
        +
            \int_0^t\,
                \Phi(z_s^x)
            \,
            ds,
            \qquad
            0\le t\le 1
\end{aligned}
\end{equation}
satisfies the uniform estimate:
\begin{equation}
\label{eq:bound}
        \|
            \Psi-\varphi
        \|_{L^\infty(\Rd; l^\infty)}
    \le
         \|\omega(\frac{d}{2n})\|_{l^\infty} 
          e^{L^V},
\end{equation}
and one gets convergence as $n \rightarrow \infty$. Moreover, $L^\Psi \le e^{L^\Phi}$ and $L^\Psi$ and $L^\Phi$ do not depend on $n$.
\end{lemma}
\begin{proof}:
As $V_j : \R^d \rightarrow \R$ for $j \in [d]$ is Lipschitz and compactly supported on $\zd$, from~\cref{HongAnas} there exists a ReLU MLP $\phi_j$ defined on $\zod$ and modulus of regularity $\omega_j$ with width at most $8d(n+1)^d$, depth at most $ \lceil log_2d \rceil + 4$ and at most $16d(n+1)^d$ free parameters such that:
\begin{equation}
\label{ReuyanResult}
    \|V_j -\phi_j\|_{L^\infty(\zod)} \leq \omega_j(\frac{d}{2n}).    
\end{equation}
Define $(\phi)_j \eqdef \phi_j$ and $( \omega)_j \eqdef  \omega_j$. $V$ is supported on a compact set in $(0,1)^d$. Without loss of generality, suppose that $V_j$ is supported on $[R_j,1-R_j]^d$ for a $R>0$.
As $\phi$ is Lipschitz on $\zod$, there exists $M \in \R$ such that:
\[
\| \phi \|_{ L^\infty( \zod ; l^\infty )} \leq M.
\]
Consider the ReLU bump function below for a small $\delta>0$ such that $[R_j-\delta, 1 -R_j +\delta] \subseteq \zod$:
\begin{equation}
\label{Eq:RelU_BumpFunct}
\begin{split}
    b_{\delta}(x)
& 
    \eqdef
    \sigma\!\left(
    1-\frac{1}{\delta}
    \sum_{i=1}^{d}
    \left[
    \sigma(R_j-x_i)
    +\sigma\bigl(x_i-(1-R_j)\bigr)
    \right]
    \right)
    \\
& 
    = 
    \begin{cases}
        0, & \text{if } x \notin [ R_j - \delta, 1 - R_j + \delta ]^d, \\
        1, & \text{if } x \in [ R_j , 1 - R_j ]^d.
    \end{cases}
\end{split}
\end{equation}
Implement the following network for $j\in [d]$:
\begin{equation}
\label{Eq:Implement_ReLU_Bumpfunc}
\begin{split}
    \mathbf{x} & \Rightarrow 
    (\phi_j, b_\delta)
    \\ 
    & \Rightarrow 
    (\phi_j, \sigma_{ReLU}(-Mb_\delta - \phi_j) , b_\delta)
    ,\\ 
    & \Rightarrow 
    (\sigma_{ReLU} \bigl(Mb_\delta - (\phi_j+ \sigma_{ReLU}(-Mb_\delta - \phi_j)) , b_\delta \bigr)
    ,\\
    & \Rightarrow 
    Mb_\delta - \sigma_{ReLU} \bigl(Mb_\delta - (\phi_j+ \sigma_{ReLU}(-Mb_\delta - \phi_j)) \bigr) = \Phi_j.
\end{split}
\end{equation}
The resulting $\Phi_j =  \min \{ Mb_\delta , \max\{ -Mb_\delta , \phi_j \}\} $ equals $\phi_j$ on $[R_j,1-R_j]^d$ and equals zero outside $[R_j-\delta,1-R_j+\delta]^d$. $\Phi_j$ has width:
\begin{align*}
    \mathrm{width}(\Phi_j) & \le \mathrm{width}(\phi_j) + 2d +1
    \\
    & \le  8d(n+1)^d + 2d +1.
\end{align*}
Its depth is:
\[
\mathrm{depth}(\Phi_j) = \mathrm{depth}(\phi_j)+ 3 = \lceil \text{log}_2d \rceil + 7.
\]
Its parameter count is:
\[
\mathrm{Par}(\Phi_j) = \mathrm{Par}(\phi_j) + 6d^2 + 9d = 16d(n+1)^d + 6d^2 + 9d.
\]
If needed, one can increase the depth of the neural networks such that $\mathrm{depth}(\Phi_j) = \max_{1 \leq i \leq d} \{ \mathrm{depth}(\Phi_i) \}$, then by parallelization~\cite[Proposition 2.3.]{petersen2024mathematical} the neural network: 
\[\Phi(x)=(\Phi_1(x), \ldots, \Phi_d(x)):\mathbb{R}^{d} \rightarrow \Rd\] has the same depth. Its width is at most: 
\[
\mathrm{width}(\Phi) 
\leq
    \sum_{j=1}^d \mathrm{width}(\Phi_j) 
\leq
    16d^2(n+1)^d + 6d^3 + 9d^2,
\]
And parameters:
\[
\mathrm{Par}(\Phi) \leq \sum_{i=1}^d \mathrm{Par}(\phi_j) + 6d^2 + 9d = 16d^2(n+1)^d + 6d^3 + 9d^2
.
\]
By \eqref{ReuyanResult} and the definition of $\Phi$ one has:
\begin{equation}
        \|\Phi - V\|_{ L^\infty( \Rd ; l^\infty )} 
    \leq
        \|\phi - V\|_{ L^\infty( \zod ; l^\infty )} 
    \le
        \|\omega(\frac{d}{2n})\|_{l^\infty} ,  
\end{equation}
this yields:
\begin{equation}
\label{ineq:Upperbound_Vectorfields}
    \begin{split}
        \|V(x_s^x)-\Phi(z_s^x)\|_{L^\infty(\Rd; l^\infty)} 
        &\leq 
            \|V(x_s^x)-V(z_s^x)\|_{L^\infty(\Rd; l^\infty)} + \| V(z_s^x)-\Phi(z_s^x)\|_{L^\infty(\Rd; l^\infty)} \\
        &\leq 
            L^{V}\| x_s^x - z_s^x \|_{L^\infty(\Rd; l^\infty)} 
        + 
            \|\omega(\frac{d}{2n})\|_{l^\infty} ,  
    \end{split}
\end{equation}
hence:
\begin{equation*}
\begin{split}
    \| z_t^x - x_t^x \|_{L^\infty(\Rd; l^\infty)} 
    &= 
        \| \int_0^t V( x_s^x ) - \Phi( z_s^x ) ds \|_{L^\infty(\Rd; l^\infty)} \\
    &\leq 
        \int_0^t \| V( x_s^x ) - \Phi( z_s^x ) \|_{L^\infty(\Rd; l^\infty)}ds \\
    &\leq 
        L^{V}\int_0^t \| x_s^x - z_s^x \|_{ L^\infty( \Rd ; l^\infty) } ds + \left( \| \omega ( \frac{d}{2n} ) \|_{ l^\infty } \right) t.
\end{split}
\end{equation*}
By applying Grönwall's inequality (\cite{Pachpatte1998} Theorem 1.3.1) 
\[
    \|z_t^x-x_t^x\|_{L^\infty(\Rd; l^\infty)} \leq \left( \|\omega(\frac{d}{2n})\|_{l^\infty} \right) te^{L^{V}t},
\]
let $t=1$ :
\begin{equation}
\label{ineq:NODE_Flow}
    \| \Psi(x)-\varphi(x)\|_{L^\infty(\Rd; l^\infty)} 
    = 
    \|z_1^x-x_1^x\|_{L^\infty(\Rd; l^\infty)}
    \leq 
        \|\omega(\frac{d}{2n})\|_{l^\infty} e^{L^{V}}.
\end{equation}
The right hand side goes to zero as $n \rightarrow \infty$. 

Observe that the Lipschitz constant $L^{\Psi}$ does \textit{not} depend on the approximation parameter $n\in \mathbb{N}_+$. This can be shown as: 
\[
\big\| z_t^x - z_t^y \big\|_{ L^\infty( \Rd ; l^\infty )}
    \le
    \big\| x-y \big\|_{ L^\infty( \Rd ; l^\infty )} + \int_0^t \big\| \Phi(z_s^x) - \Phi(z_s^y) \big\|_{ L^\infty( \Rd ; l^\infty )}ds
        \le 
        \big\| x-y \big\|_{ L^\infty( \Rd ; l^\infty )} + L^{\Phi}\int_0^t \big\| z_s^x - z_s^y \big\|_{ L^\infty( \Rd ; l^\infty )}ds.
\]
By applying Grönwall's inequality \cite[Theorem 1.2.2]{Pachpatte1998} 
\[
    \|z_t^x-z_t^y\|_{L^\infty(\Rd; l^\infty)} \leq \|x-y\|_{L^\infty(\Rd; l^\infty)} e^{tL^{\Phi}}.
\]
Let $t=1$ :
\begin{equation}
\label{ineq:NODE}
    \| \Psi(x)-\Psi(y)\|_{L^\infty(\Rd; l^\infty)} 
    = 
    \|z_1^x-z_1^y\|_{L^\infty(\Rd; l^\infty)} \leq \|x-y\|_{L^\infty(\Rd; l^\infty)}e^{L^{\Phi}}.
\end{equation}
This means that $L^\Psi \leq e^{L^\Phi} $. Since the Lipschitz constant of $\phi$ and the function $b_\delta$ do not depend on $n$, $L^\Phi$ is also independent of $n$. As a result, $L^\Psi$ is independent of $n$ and the result follows.   
\end{proof}

\begin{proposition}[Universal Approximation by Deep Neural ODEs]
\label[proposition]{thrm:Universality_of_Neural_ODEs_RH}
Let
$\varphi\in \Homeo[\Rd][\zd][I]$ and $n,d, I \in \N_+$
then, 
there exist ReLU Neural ODEs $\{\Psi_i=\mathrm{Flow}(\Phi_i)\}_{i=1}^{I}$ such that $ \Psi \eqdef \bigcirc_{i=1}^{I}\,\Psi_i$ satisfies the approximation guarantee:
\begin{equation}
\label{eq:Universality_of_Neural_ODEs}
         \big\| \varphi - \Psi \big\|_{L^\infty(\Rd; l^\infty)}
    \le
     \sum_{i=1}^{I} 
        \left( \|\omega_i(\frac{d}{2n})\|_{l^\infty}
        \prod_{j=i}^{I} e^{L^{V_j}}  \right),
\end{equation}
where $(\omega_i)_j$ is the modulus of regularity of $(V_i)_j$ for $i \in [I]$ and $j \in [d]$. The right hand side converges to zero as $n\rightarrow \infty$. \\
Moreover, $\Psi \in \Homeo[\Rd][\zd]$, with Lipschitz constant at most $\prod_{i=1}^{I} e^{L^{\Phi_i}}$ and $\Phi_1,\dots,\Phi_I$ are ReLU MLP vector fields of depth at most $\lceil \text{log}_2d \rceil + 7$, width at most $8d^2(n+1)^d + 2d^2 + d$, and at most $16d^2(n+1)^d + 6d^3 + 9d^2$ parameters. Furthermore, $L^{\Phi_i}$ and $L^\Psi$ do not depend on the parameter $n$ for $i \in [I]$. \\ 
\end{proposition}
\begin{proof}:
Recall \eqref{eq:defn:HomeoStructure__Representation_Homeo} as $\varphi\in \Homeo[\Rd][\zod][I]$ there exist $\varphi_i = \Flow(V_i)$ for every $i\in [I]$ such that:
\begin{equation}
\label{eq:Compos_varphi__RH}
    \varphi = \bigcirc_{i=1}^{I} \, \varphi_i(x).
\end{equation}
We use \cref{lem:Vectorfield_to_Flow} for each $\varphi_i$ to find Neural ODEs $\{\Psi_i\}_{i=1}^I$.

Define $\Psi \eqdef \bigcirc_{i=1}^I \Psi_i$. We try to show that the Lipschitz constant $L^{\Psi}$ is at most $\prod_{i=1}^{I} e^{L^{\Phi_i}}$ using induction. The base case is already proved in \cref{lem:Vectorfield_to_Flow}; for the step of the induction, consider the induction assumption below :
\[
\big\| \bigcirc_{i=1}^{I-1}\,\Psi_i(x)-\bigcirc_{i=1}^{I-1}\,\Psi_i(y) \big\|_{L^\infty(\Rd; l^\infty)}
        \le 
        \small\prod_{i=1}^{I-1} e^{L^{\Phi_i}}\big\| x-y \big\|_{L^\infty(\Rd; l^\infty)},
\]
we can write:
\begin{equation}
\label{ineq:DeepNeuralODE___DistinctFlow}
\big\| \bigcirc_{i=1}^{I}\,\Psi_i(x)-\bigcirc_{i=1}^{I}\,\Psi_i(y) \big\|_{L^\infty(\Rd; l^\infty)}
    \le 
    e^{L^{\Phi_I}}
    \big\| \bigcirc_{i=1}^{I-1}\,\Psi_i(x)-\bigcirc_{i=1}^{I-1}\,\Psi_i(y) \big\|_{L^\infty(\Rd; l^\infty)}
        \le 
        \small\prod_{i=1}^{I} e^{L^{\Phi_i}}\big\| x-y \big\|_{L^\infty(\Rd; l^\infty)},
\end{equation}
So one can deduce the Lipschitz constant is at most 
$\small\prod_{i=1}^{I} e^{L^{\Phi_i}}$ and by \cref{lem:Vectorfield_to_Flow} it doesn't depend on $n$. \\

We again use induction to show inequality \eqref{eq:Universality_of_Neural_ODEs}. The base case is already proved in  \cref{lem:Vectorfield_to_Flow}. For the step of the induction, consider the induction assumption below:
\[
    \big\| \bigcirc_{i=1}^{I-1} \varphi_i
        -
            \bigcirc_{i=1}^{I-1} \, \Psi_i \big\|_{L^\infty(\Rd; l^\infty)} 
            \leq
                \sum_{i=1}^{I-1}
                    \left( \|\omega_i(\frac{d}{2n})\|_{l^\infty}
                    \prod_{j=i}^{I-1} e^{L^{V_j}}  \right) .
\]
Using \eqref{eq:Compos_varphi__RH} and \cref{lem:Vectorfield_to_Flow} one can write:
\begin{equation*}
    \begin{split}
        \big\| \varphi - \bigcirc_{i=1}^{I} \, \Psi_i \big\|_{L^\infty(\Rd; l^\infty)} 
        &=
        \big\| \bigcirc_{i=1}^{I}\varphi_i - \bigcirc_{i=1}^{I} \, \Psi_i \big\|_{L^\infty(\Rd; l^\infty)}  \\
        &   \le
            \big\| \bigcirc_{i=1}^{I} \, \varphi_i 
            - 
            \varphi_I \circ \bigcirc_{i=1}^{I-1} \Psi_i \big\|_{L^\infty(\Rd; l^\infty)}  \\
            &\quad \quad+
            \big\|\varphi_I \circ \bigcirc_{i=1}^{I-1} \Psi_i 
            -
                \bigcirc_{i=1}^{I} \, \Psi_i \big\|_{L^\infty(\Rd; l^\infty)}  \\
            & \le
                e^{L^{V_I}} 
                \big\| \bigcirc_{i=1}^{I-1} \varphi_i 
                -
                \bigcirc_{i=1}^{I-1} \, \Psi_i \big\|_{L^\infty(\Rd; l^\infty)} 
                +
                e^{L^{V_I}}
                \|\omega_I(\frac{d}{2n})\|_{l^\infty} .
    \end{split}
\end{equation*}
Then by the induction assumption:
\[
\big\| \varphi - \bigcirc_{i=1}^{I} \, \Psi_i \big\|_{L^\infty(\Rd; l^\infty)} 
    \le
    \sum_{i=1}^{I} 
        \left( \|\omega_i(\frac{d}{2n})\|_{l^\infty}
        \prod_{j=i}^{I} e^{L^{V_j}}  \right).
\]
By \cref{lem:Vectorfield_to_Flow} the right hand side goes to zero as $n\rightarrow \infty$. 
\end{proof}
\subsubsection{Dependent Lipschitz Constant for Neural ODE, Better Rates}
Naturally, one may wonder if improved rates are achievable under additional smoothness of the target homeomorphism.
\begin{lemma}[Transfer: Universal Approximation of Vector Field to Flow (Differentiable Case)]
\label[lemma]{lem:Vectorfield_to_Flow_SCase}
Fix $k,d,L\in \mathbb{N}_+$ and $N_j \in \N_+$ for $j \in [d]$, let $V= (V_1,\dots,V_d)\in \mathcal{X}^k(\Rd,\zd)$. Define for $j \in [d]$ , $\upsilon_j \eqdef{85 (k+1)^d 8^k \|V_j\|_{C^k([0,1]^d)}({N_j} {L})^{-2k/d}}$  and $\upsilon(N_1,\dots,N_d,L) \eqdef (\upsilon_1,\dots,\upsilon_d)$; then for every time-one flow:
\begin{equation}
\label{Eq:timeonedlowof_V}
\begin{aligned}
        \varphi(x)
    & =
        x_1^x
\\
        x_t^x
    &=
            x
        +
            \int_0^t\,
                V(x_s^x)
            \,
            ds,
            \qquad
            0\le t\le 1,
\end{aligned}
\end{equation}
there exists a ReLU MLP \(\Phi:\mathbb{R}^{d} \rightarrow \Rd,\)
compactly supported on $(0,1)^d$ with width at most $\sum_{j=1}^d 1 +2d + 17 k d^{\,d+13} d(N_j+2) \log_2(8N_j)$ and depth at most $18k^2 (L+2) \log_2(4L) + 2d + 3 $
such that the Neural ODE:
\begin{equation}
\label{eq_lem:decomposition__flow_representation___MLPVersion}
\begin{aligned}
        \Psi(x)
    & =
        z_1^x,
\\
        z_t^x
    &=
            x
        +
            \int_0^t\,
                \Phi(z_s^x)
            \,
            ds,
            \qquad
            0\le t\le 1, 
\end{aligned}
\end{equation}
satisfies the uniform estimate:
\begin{equation}
\label{eq:bound__2}
        \|
            \Psi-\varphi\|_{L^\infty(\R^d; l^\infty)}
    \le
        \|\upsilon(N_1,\dots,N_d,L)\|_{l^\infty}
        e^{L^{V}}.
\end{equation}
In terms of number of parameters $P$ one can write:
\begin{equation}
    \| \Psi-\varphi\|_{L^\infty( \Rd ; l^\infty )} 
    \in
        \mathcal{O} \big( (\frac{P}{\log P})^{-2k/d} \big).
\end{equation}
Note that $\Psi\in \Homeo[\Rd][\zd]$.
\end{lemma}
\begin{proof}:
For any $N_j, L \in \mathbb{N}_+$ by~\citep[Theorem 1.1]{lu2021deep} for $V_j $, $j \in [d]$ 
there exists a function $\phi_j$ defined on $\zod$, implemented by a ReLU FNN with width at most $C_1 (N_j+2) \log_2(8N_j)$  
and depth at most $C_2 (L+2) \log_2(4L) + 2d$ such that:
\begin{equation}
\label{ineq_Lu}
\|\phi_j - V_j\|_{L^\infty([0,1]^d)} \le C_3 \|V_j\|_{C^k([0,1]^d)}({N_j} {L})^{-2k/d},
\end{equation}
where $C_1 = 17 k^{d+1}3^dd$, $C_2 = 18 k^2$, and $C_3 = 85 (k+1)^d 8^k$. 
Define $(\phi)_j \eqdef \phi_j$.
$V$ is supported on a compact set in $(0,1)^d$. Without loss of generality, suppose that $V_j$ is supported on $[R,1-R]^d$ for a $R>0$.
As $\phi$ is Lipschitz on $\zod$, there exists $M \in \R$ such that:
\[
\| \phi \|_{ L^\infty( \zod ; l^\infty )} \leq M.
\]
Consider the ReLU bump function below for a small $\delta>0$ such that $[R-\delta, 1 -R +\delta] \subseteq \zod$:
\begin{equation}
\label{Eq:RelU_BumpFunct}
\begin{split}
    b_{R,\delta}(x)
& 
    \eqdef
    \sigma\!\left(
    1-\frac{1}{\delta}
    \sum_{i=1}^{d}
    \left[
    \sigma(R-x_i)
    +\sigma\bigl(x_i-(1-R)\bigr)
    \right]
    \right)
    \\
& 
    = 
    \begin{cases}
        0, & \text{if } x \notin [ R - \delta, 1 - R + \delta ]^d, \\
        1, & \text{if } x \in [ R , 1 - R ]^d.
    \end{cases}
\end{split}
\end{equation}
Implement the following network for $j\in [d]$:
\begin{equation}
\label{Eq:Implement_ReLU_Bumpfunc}
\begin{split}
    \mathbf{x} & \Rightarrow 
    (\phi_j, b_\delta)
    \\ 
    & \Rightarrow 
    (\phi_j, \sigma_{ReLU}(-Mb_\delta - \phi_j) , b_\delta)
    ,\\ 
    & \Rightarrow 
    (\sigma_{ReLU} \bigl(Mb_\delta - (\phi_j+ \sigma_{ReLU}(-Mb_\delta - \phi_j)) , b_\delta \bigr)
    ,\\
    & \Rightarrow 
    Mb_\delta - \sigma_{ReLU} \bigl(Mb_\delta - (\phi_j+ \sigma_{ReLU}(-Mb_\delta - \phi_j)) \bigr) = \Phi_j.
\end{split}
\end{equation}
The resulting $\Phi_j =  \min \{ Mb_\delta , \max\{ -Mb_\delta , \phi_j \}\} $ equals $\phi_j$ on $[R,1-R]^d$ and equals zero outside $[R-\delta,1-R+\delta]^d$. $\Phi_j$ has width:
\begin{align*}
    \mathrm{width}(\Phi_j) & \le \mathrm{width}(\phi_j) + 2d + 1
    \\
    & \le  C_1 (N_j+2) \log_2(8N_j) + 2d + 1
    \\
    & \le 17 k^{d+1}3^dd (N_j+2) \log_2(8N_j) + 2d + 1.
\end{align*}
Its depth is:
\[
\mathrm{depth}(\Phi_j) = \mathrm{depth}(\phi_j) + 3 = C_2 (L+2) \log_2(4L) + 2d + 3.
\]
If needed, one can increase the depth of the neural networks such that $\mathrm{depth}(\Phi_j) = \max_{1 \leq i \leq d} \{ \mathrm{depth}(\Phi_i) \}$, then by parallelization~\cite[Proposition 2.3.]{petersen2024mathematical} the neural network: 
\[\Phi(x)=(\Phi_1(x), \ldots, \Phi_d(x)):\mathbb{R}^{d} \rightarrow \Rd\] has the same depth. Its width is at most: 
\[
\mathrm{width}(\Phi) \leq \sum_{j=1}^d \mathrm{width}(\Phi_j),
\]
By \eqref{ineq_Lu} and the definition of $\Phi$ one has:
\begin{equation}
        \|\Phi - V\|_{ L^\infty( \Rd ; l^\infty )} 
    \leq
        \|\phi - V\|_{ L^\infty( \zod ; l^\infty )} 
    \le
        \| \upsilon(N_1,\dots,N_d,L) \|_{l^\infty},
\end{equation}
this yields:
\begin{equation}
\begin{split}
    \|V(x_s^x)-\Phi(z_s^x)\|_{L^\infty(\Rd; l^\infty)} 
        &\leq 
            \|V(x_s^x)-V(z_s^x)\|_{L^\infty(\Rd; l^\infty)} + \| V(z_s^x)-\Phi(z_s^x)\|_{L^\infty(\Rd; l^\infty)} \\
        &\leq 
            L^{V}\| x_s^x - z_s^x \|_{L^\infty( \Rd ; l^\infty)} 
        + 
            \|\upsilon(N_1,\dots,N_d,L)\|_{l^\infty}.
\end{split}
\end{equation}
Then:
\begin{equation}
\label{enq_pre_gronwall__v0}
\begin{split}
\|x_t^x-z_t^x\|_{L^\infty(\Rd; l^\infty)} 
&= \|\int_0^tV(x_s^x)-\Phi(z_s^x)ds\|_{L^\infty(\Rd; l^\infty)} \\
    & \leq \int_0^t\|V(x_s^x)-\Phi(z_s^x)\|_{L^\infty(\Rd; l^\infty)}ds \\
        &\leq L^V\int_0^t\|x_s^x-z_s^x\|_{L^\infty(\Rd; l^\infty)}ds + \left 
        (
        \|\upsilon(N_1,\dots,N_d,L)\|_{l^\infty}
        \right) t.
\end{split}
\end{equation}
By applying Grönwall's inequality \cite[Theorem 1.3.1]{Pachpatte1998} 
\[
    \|z_t^x-x_t^x\|_{L^\infty(\Rd; l^\infty)} \leq    \|\upsilon(N_1,\dots,N_d,L)\|_{l^\infty} t e^{L^Vt}.
\]
Let $t=1$ :
\begin{equation*}
\label{ineq:NODE_Flow__2}
    \| \Psi(x)-\varphi(x)\|_{L^\infty( \Rd ; l^\infty )} 
    = 
    \|z_1^x-x_1^x\|_{L^\infty( \Rd ; l^\infty )} 
        \leq 
        \|\upsilon(N_1,\dots,N_d,L)\|_{l^\infty}
        e^{L^V}
.
\end{equation*}
By~\citep[Corollary 1.3]{lu2021deep} one can express the rate in terms of the parameter count $P$ as:
\begin{equation}
    \| \Psi-\varphi\|_{L^\infty( \Rd ; l^\infty )} 
    \in
        \mathcal{O} \big( (\frac{P}{\log P})^{-2k/d} \big).
\end{equation}
To achieve the Lipschitz constant, as $\Phi$ is Lipschitz one can write:
\[
\big\| z_t^x - z_t^y \big\|_{ L^\infty( \Rd ; l^\infty )}
    \le
    \big\| x-y \big\|_{ L^\infty( \Rd ; l^\infty )} + \int_0^t \big\| \Phi(z_s^x) - \Phi(z_s^y) \big\|_{ L^\infty( \Rd ; l^\infty )}ds
        \le 
        \big\| x-y \big\|_{ L^\infty( \Rd ; l^\infty )} + L^{\Phi}\int_0^t \big\| z_s^x - z_s^y \big\|_{ L^\infty( \Rd ; l^\infty )}ds.
\]
By applying Grönwall's inequality \cite[Theorem 1.2.2]{Pachpatte1998} 
\[
    \|z_t^x-z_t^y\|_{ L^\infty( \Rd ; l^\infty )} \leq \left(\|x-y\|_{ L^\infty( \Rd ; l^\infty )} \right)e^{L^{\Phi}t}.
\]
Set $t=1$:
\begin{equation}
\label{ineq:NODE__3}
    \big\| \Psi(x) - \Psi(y) \big\|_{ L^\infty( \Rd ; l^\infty )}
    = \big\| z_1^x - z_1^y \big\|_{ L^\infty( \Rd ; l^\infty )}
    \le
    \left(\|x-y\|_{ L^\infty( \Rd ; l^\infty )} \right)e^{L^{\Phi}}.
\end{equation}
The result follows.
\end{proof}

\begin{proposition}[Universal Approximation by Deep Neural ODEs (Differentiable Case)]
\label[proposition]{prop:Universality_of_Neural_ODEs}
Consider $k, d, I , L \in \N_+$ and $N_j \in \N_+$ for $j \in [d]$ and $\varphi\in \Diff[\Rd][\zd][k,I]$; then, there exist ReLU Neural ODEs $\{\Psi_i=\mathrm{Flow}(\Phi_i)\}_{i=1}^{I}$ such that $ \Psi \eqdef \bigcirc_{i=1}^{I}\,\Psi_i$ satisfies the approximation guarantee:
\begin{equation}
\label{eq:Universality_of_Neural_ODEs__2_}
\begin{split}
\big\| \varphi - \Psi \big\|_{L^\infty(\R^d; l^\infty)} 
\le
    \sum_{i=1}^{I} \left( \|\upsilon_i(N_1,\dots,N_d,L)\|_{l^\infty} 
        \prod_{j=i}^{I} e^{L^{V_j}}  \right) .
\end{split}
\end{equation}
Where $(\upsilon_i)_j = 85 (k+1)^d 8^k \|(V_i)_j\|_{C^k([0,1]^d)}({N_j} {L})^{-2k/d}$ for $i \in [I]$ and $j \in [d]$. The right hand side goes to zero as $L\rightarrow \infty$ and $N_j \rightarrow \infty$ for each $j \in [d]$. \\
Moreover, $\Psi \in \Homeo[\Rd][\zd]$, with Lipschitz constant at most $\prod_{i=1}^{I} e^{L^{\Phi_i}}$ where $\Phi_i$ are ReLU MLPs of width at most $\sum_{j=1}^d 1 + 2d + 17 k^{d+1}3^dd(N_j+2) \log_2(8N_j)$ and depth at most $18k^2 (L+2) \log_2(4L) + 2d + 3 $ and $N_j,L\in \N_+$ for each $j \in [d]$ , $i \in [I]$. Note that $L^\Psi$ depends on the parameters $N_1,\dots,N_d$. Furthermore, if $L$ is not fixed, then in terms of the number of parameters $P$ one can write:
\begin{equation}
    \| \Psi-\varphi\|_{L^\infty( \Rd ; l^\infty )} 
    \in
        \mathcal{O} \big( (\frac{P}{\log P})^{-2k/d} \big).
\end{equation}
\end{proposition}
\begin{proof}:
Recall as in \eqref{eq:defn:HomeoStructure__Representation}, $\varphi\in \Diff[\Rd][\zd][k,I]$  means $\varphi_i = \Flow(V_i)$ for every $i \in [I]$ and :
\begin{equation}
\label{eq:Compos_varphi__RH2}
    \varphi = \bigcirc_{i=1}^{I} \, \varphi_i(x).
\end{equation}
We use \cref{lem:Vectorfield_to_Flow_SCase} for each $\varphi_i$ to find Neural ODEs $\{\Psi_i=\mathrm{Flow}(\Phi_i)\}_{i=1}^I$.

Define $\Psi \eqdef \bigcirc_{i=1}^I \Psi_i$. We try to show that the Lipschitz constant $L^{\Psi}$ is at most $\prod_{i=1}^{I} e^{L^{\Phi_i}}$ using induction. The base case is already proved in \cref{lem:Vectorfield_to_Flow_SCase}; for the step of the induction, consider the induction assumption below :
\[
\big\| \bigcirc_{i=1}^{I-1}\,\Psi_i(x)-\bigcirc_{i=1}^{I-1}\,\Psi_i(y) \big\|_{ L^\infty( \Rd ; l^\infty )}
        \le 
        \small\prod_{i=1}^{I-1} e^{L^{\Phi_i}}\big\| x-y \big\|_{ L^\infty( \Rd ; l^\infty )}.
\]
Note:
\begin{equation}
\label{ineq:DeepNeuralODE___DistinctFlow}
\big\| \bigcirc_{i=1}^{I}\,\Psi_i(x)-\bigcirc_{i=1}^{I}\,\Psi_i(y) \big\|_{ L^\infty( \Rd ; l^\infty )}
    \le 
    e^{L^{\Phi_I}}
    \big\| \bigcirc_{i=1}^{I-1}\,\Psi_i(x)-\bigcirc_{i=1}^{I-1}\,\Psi_i(y) \big\|_{ L^\infty( \Rd ; l^\infty )}
        \le 
        \small\prod_{i=1}^{I} e^{L^{\Phi_i}}\big\| x-y \big\|_{ L^\infty( \Rd ; l^\infty )}.
\end{equation}
So one can deduce the Lipschitz constant is at most 
$\small\prod_{i=1}^{I} e^{L^{\Phi_i}}$. \\
We again use induction to show inequality \eqref{eq:Universality_of_Neural_ODEs__2_}. The base case is already proved in \cref{lem:Vectorfield_to_Flow_SCase}. For the step of the induction consider the induction assumption below:
\begin{equation*}
\big\| \bigcirc_{i=1}^{I-1} \varphi_i
    -
        \bigcirc_{i=1}^{I-1} \, \Psi_i \big\|_{L^\infty(\Rd; l^\infty)}
        \leq 
            \sum_{i=1}^{I-1} \left( \|\upsilon_i(N_1,\dots,N_d,L)\|_{l^\infty} 
            \prod_{j=i}^{I-1} e^{L^{V_j}}  \right)          .
\end{equation*}
Using \eqref{eq:Compos_varphi__RH2} one can write:
\begin{equation*}
    \begin{split}
         \big\| \varphi - \bigcirc_{i=1}^{I} \, \Psi_i \big\|_{L^\infty(\Rd; l^\infty)}
        &=
        \big\| \bigcirc_{i=1}^{I}\varphi_i - \bigcirc_{i=1}^{I} \, \Psi_i \big\|_{L^\infty(\Rd; l^\infty)} \\
        &   \le
            \big\| \bigcirc_{i=1}^{I} \, \varphi_i
            - 
            \varphi_I \circ \bigcirc_{i=1}^{I-1} \Psi_i \big\|_{L^\infty(\Rd; l^\infty)} \\
            &\quad \quad+
            \big\|\varphi_I \circ \bigcirc_{i=1}^{I-1} \Psi_i 
            -
                \bigcirc_{i=1}^{I} \, \Psi_i \big\|_{L^\infty(\Rd; l^\infty)} \\
            & \le 
            \big\| \bigcirc_{i=1}^{I-1} \varphi_i 
                -
                    \bigcirc_{i=1}^{I-1} \, \Psi_i \big\|_{L^\infty(\Rd; l^\infty)} e^{L^{V_I}} \\
            &\quad \quad+
                \|\upsilon_I(N_1,\dots,N_d,L)\|_{l^\infty} e^{L^{V_I}}.
    \end{split}
\end{equation*}
Then, by the induction assumption:
\[
\big\| \varphi - \bigcirc_{i=1}^{I} \, \Psi_i \big\|_{L^\infty(\Rd; l^\infty)}
    \le
        \sum_{i=1}^{I} \left( \|\upsilon_i(N_1,\dots,N_d,L)\|_{l^\infty} 
        \prod_{j=i}^{I} e^{L^{V_j}}  \right) .
\]
By \cref{lem:Vectorfield_to_Flow_SCase} the right hand side goes to zero as $L \rightarrow \infty$ and $N_j\rightarrow \infty$ for each $j \in [d]$.
In terms of the number of parameters $P$, one can write from \cref{lem:Vectorfield_to_Flow_SCase}:
\begin{equation}
    \big\| \varphi - \bigcirc_{i=1}^{I} \, \Psi_i \big\|_{L^\infty(\Rd; l^\infty)}
    \in
        \mathcal{O} \big( (\frac{P}{\log P})^{-2k/d} \big).
\end{equation}
We have thus concluded the proof.
\end{proof}
\begin{proposition}[Universal Approximation by Deep Neural ODEs (Differentiable Case)]
\label[proposition]{prop:Universality_of_Neural_ODEs__DifableCase}
Let $\varphi\in \Diff[\Rd][\zd][k,I]$ and $N,\,L,\,k \in \N_+$
then, 
there exist ReLU Neural ODEs $\{\Psi_i=\mathrm{Flow}(\Phi_i)\}_{i=1}^{I}$ such that $ \Psi \eqdef \bigcirc_{i=1}^{I}\,\Psi_i$ satisfies the approximation guarantee:
\begin{equation}
\label{eq:Universality_of_Neural_ODEs__2___}
\begin{split}
\big\| \varphi - \Psi \big\|_{L^\infty(\R^d; l^\infty)} \le
    \sum_{i=1}^{I} \left( \|\upsilon_i(N,L)\|_{l^\infty} 
        \prod_{j=i}^{I} e^{L^{V_j}}  \right).
\end{split}
\end{equation}
Where $(\upsilon_i)_j = 85 (k+1)^d 8^k \|(V_i)_j\|_{C^k([0,1]^d)}(NL)^{-2k/d}$ and the right hand side goes to zero as $L\rightarrow \infty$ and $N \rightarrow \infty$.\\ 
Moreover, $\Psi \in \Homeo[\Rd][\zd]$ and its Lipschitz constant is at most $\prod_{i=1}^{I} e^{L^{\Phi_i}}$. $\Phi_1,\dots,\Phi_i$ are ReLU MLPs of width at most $ d + 2d^2 + 17 k^{d+1}3^dd^2(N+2) \log_2(8N)$ and depth at most $18k^2 (L+2) \log_2(4L) + 2d + 3 $. Note that $L^\Psi$ depends on the parameter $N$. Furthermore, in terms of the number of parameters $P$, one can write:
\begin{equation}
    \| \Psi-\varphi\|_{L^\infty( \Rd ; l^\infty )} 
    \in
        \mathcal{O} \big( (\frac{P}{\log P})^{-2k/d} \big).
\end{equation}
\end{proposition}
\begin{proof}:
    Set $N = N_1= N_2= \dots = N_d$ then the result follows from \cref{prop:Universality_of_Neural_ODEs}. 
\end{proof}
\begin{table}[H]
\centering
\begin{tabular}{lll}
\toprule
\textbf{Complexity of MLP Vector fields} & \textbf{Lipschitz Case} & \textbf{$C^k$-Differentiable Case ($k \ge 1$)}\\
\midrule
Depth & $\lceil \log_2(d) \rceil + 7$ &  $18k^2 (L+2) \log_2(4L) + 2d+ 3 $\\
Width & $d + 2d^2 + 8d^2(n+1)^d$ & $d + 2d^2 + 17 k^{d+1}3^dd^2(N+2) \log_2(8N)$ \\
Nonzero parameters & $16d^2(n+1)^d + 6d^3 + 9d^2$ & Explicit count not available \\
\bottomrule
\end{tabular}
\caption{Parametric Complexity of the ReLU MLP vector fields used in \cref{thrm:Universality_of_Neural_ODEs_RH} and \cref{prop:Universality_of_Neural_ODEs__DifableCase}.}
\label{Table_Complexity}
\end{table}
\subsubsection{Decomposition}
\label{s:Proof__ss:Step_1}
\begin{lemma}[Normality]
\label[lemma]{lem:normality}
For an open set $K \subseteq M$, the set $F(K)$ generates a normal subgroup (denoted as) $\langle F(K)  \rangle$ of $\Diffk[K][\infty]$.
\end{lemma}
\begin{proof}:

Let  $F \in \Diffk[K][\infty]$ and $\varphi \in F(K)$ so there exists $V\in \mathcal{X}^\infty(K)$ such that $\varphi = \Flow(V)$, then by \citep[Corollary 9.14]{Lee_2013__SmoothManifolds} the time-one flow of the (pushforward) vector field $F_*V \in \mathcal{X}^\infty(K)$ is $\eta = F \circ \varphi \circ F^{-1}$, in other words $\eta = \Flow(F_*V)$ which means $\eta \in F(K)$. 

Now let $\varphi \in \langle F(K) \rangle$. This means there exists $n \in \N_+$ such that one can write $\varphi$ as a composition of $n$ time-one flows:
\[ 
        \varphi =  \varphi_n\circ \dots \circ \varphi_1 .
\]
Now if $F \in \Diffk[K][\infty]$, one can write:
\[
F \circ \varphi \circ F^{-1} =
    (F \circ \varphi_n \circ F^{-1}) \circ 
        (F \circ \varphi_{n-1} \circ F^{-1}) 
        \circ \dots \circ 
            (F \circ \varphi_1 \circ F^{-1} )
    = 
    \eta_n \circ \dots \circ \eta_1 \in \langle F(K) \rangle.
\]
\end{proof} 
\begin{lemma}[Decomposition]
\label[lemma]{lem:decomposition}
Let $K \subseteq M$ where $K$ is open and connected. Then for every diffeomorphism $\varphi\in \Diff[M][K][\infty]$, there exist $I \in \N_+$ and flows $\varphi_1,\dots,\varphi_i \in F(M,K)$ such that:
\begin{equation}
\label{eq_lem:decomposition}
    \varphi 
= 
    \bigcirc_{i=1}^{I}\, 
        \varphi_i
\eqdef
    \varphi_I \circ \dots \circ \varphi_1
.
\end{equation}
\end{lemma}
\begin{proof}:
By Thurston's Theorem, see e.g.~\citep[Theorem 2.1.1]{Banyaga_1997__DiffeoBook}, the group $\Diff[K][K][\infty]$ is simple; i.e.\ it has no \textit{proper} normal subgroups besides the trivial group consisting only of the identity. Since $\langle F(K,K) \rangle \neq \{\mathrm{id}_M\}$ and by Lemma~\eqref{lem:normality}, $\langle F(K,K) \rangle $ is a normal subgroup of $ \Diff[K][K][\infty]$, which means $F(K,K)$ generates the entire group $\Diff[K][K][\infty]$.  
Now for any $\varphi \in \Diff[M][K][\infty]$, one has $\varphi_{|K} \in \Diff[K][K][\infty]$ which means:
\begin{equation}
\label{Representation_With_Flow}
    \varphi_{|K} = \bigcirc_{i=1}^I \varphi_i, \qquad \varphi_i = \Flow(V_i).
\end{equation}
As $V_i$ are compactly supported on the open set $K$ one can extend each $V_i$ by zero outside $K$.  Note that $\varphi = \mathrm{id}$ outside $K$, so considering the composition of the time-one flows of the extended vector fields finishes the proof.
\end{proof}

\subsubsection{The uniform bound}

\paragraph{Conjugation-invariant norm:} 
    A conjugation-invariant norm $\nu : G \to [0; +\infty)$ on a group $G$ is a function which satisfies the following axioms:
        \begin{itemize}
            \item[(i)] $\nu(1) = 0,$
            \item[(ii)] $\nu(f) = \nu(f^{-1}), \quad \forall f \in G;$
            \item[(iii)] $\nu(fg) \leq \nu(f) + \nu(g), \quad \forall f, g \in G;$
            \item[(iv)] $\nu(f) = \nu(gfg^{-1}), \quad \forall f, g \in G;$
            \item[(v)] $\nu(f)>0,$ for all $f\neq1$
        \end{itemize}
        
\paragraph{BIP-meagre groups:}
\label{MeagerGroup_Definition}
A group \(G\) is called BIP-meagre if every
conjugation-invariant norm on \(G\) is bounded and discrete.

\paragraph{Portable manifold:} 
\label{Potable_def}
We say that a smooth connected open manifold \(M\) is portable if it
admits a complete vector field \(X\) and a compact subset \(M_0\) with
the following properties:
\begin{itemize}
    \item \(M_0\) is an attractor of the flow \(x_t\) generated by \(X\):
    for every compact subset \(K\subset M\), there exists \(T>0\) such
    that
    \[
    x_T(K)\subset M_0.
    \]

    \item There exists a diffeomorphism
    \( f \in \Diff[M][M] \) such that
    \[
    f(M_0)\cap M_0=\varnothing.
    \]
\end{itemize}
The set \(M_0\) is called the core of a portable manifold \(M\).
\begin{lemma}
\label[lemma]{lem:Finite_ConjInv_Norm}
The group \(\Diff[M][M][\infty]\) is \hyperref[MeagerGroup_Definition]{BIP-meagre} provided that the smooth connected open manifold \(M\) is
\hyperref[Potable_def]{portable}.
\end{lemma}
\begin{proof}:
    For the proof, see \cite[Theorem 1.17]{BuragoIvanovPolterovich2008}.
\end{proof}
\begin{lemma}
\label[lemma]{thrm:structure}
There exists $I_d\in \mathbb{N}_+$ such that for any integer $I>I_d$, $\Diff[\Rd][\zd][\infty,I]$ is empty.
\end{lemma}
\begin{proof}:
Define $S \eqdef \{\, \Flow(V) \mid V \in \calX^\infty(\zd,\zd) \,\}$ and the norm: 
$$\|f\|_{\mathrm{frag}} =
\begin{cases}
0, & f = \mathrm{id}, \\[6pt]
\displaystyle \min\{\, m \in \mathbb{N} \mid f = h_1 \cdots h_m,\; h_i \in S \text{ for } i \in [m] \,\}, & f \neq \mathrm{id}.
\end{cases}$$
One can check that this norm is conjugation-invariant.
As $\zd$ is \hyperref[Potable_def]{portable}, (it is diffeomorphic to an open ball), 
by \cref{lem:Finite_ConjInv_Norm} the autonomous norm on the group $\Diff[\zd][\zd][\infty]$ is bounded by a constant $I_d$. As for any $\varphi \in \Diff[\Rd][\zd][\infty]$ one has $\varphi_{|\zd} \in \Diff[\zd][\zd][\infty]$, this means any $\varphi \in \Diff[\Rd][\zd][\infty]$ has a minimal representation less than or equal to $I_d$.
\end{proof}

\section{Positive Result Propositions and Lemmas for the Cube Case}
\label{App:ZOD_Positive}
\subsection{Independent Lipschitz Constant for Neural ODE, but Worse Rates}
\label{s:ZOD_Proof__ss:Step_2}
\begin{lemma}[Transfer: Universal Approximation of Vector Field to Flow]
\label[lemma]{lem:ZOD_Vectorfield_to_Flow}
Let $n\in \N_+$ and $V = (V_1 , \dots , V_d) \in \calX^0(\zod)$ with
modulus of regularity \(\omega_j:[0,d]\to\mathbb{R}\) for $j \in [d]$. Consider its time-one flow:
\begin{equation}
\label{Eq:ZOD_timeonedlowof_V}
\begin{aligned}
        \varphi(x)
    & =
        x_1^x
\\
        x_t^x
    &=
            x
        +
            \int_0^t\,
                V(x_s^x)
            \,
            ds,
            \qquad
            0\le t\le 1,
\end{aligned}
\end{equation}
then there exists a ReLU MLP $\Phi: \zod \to \Rd$ tangent to the faces of the cube with width at most $8d^2(n+1)^d$, depth at most $ \lceil log_2d \rceil + 4$ and at most $16d^2(n+1)^d$ free parameters such that the time-one flow (Neural ODE):
\begin{equation}
\label{eq_lem:ZOD_decomposition__flow_representation___MLPVersion___RH}
\begin{aligned}
        \Psi(x)
    & =
        z_1^x,
\\
        z_t^x
    &=
            x
        +
            \int_0^t\,
                \Phi(z_s^x)
            \,
            ds,
            \qquad
            0\le t\le 1
\end{aligned}
\end{equation}
satisfies the uniform estimate:
\begin{equation}
\label{eq:ZOD_bound}
        \|
            \Psi-\varphi
        \|_{L^\infty( \zod ; l^\infty)}
    \le
         \|\omega(\frac{d}{2n})\|_{l^\infty} 
          e^{L^V},
\end{equation}
and one gets convergence as $n \rightarrow \infty$. Moreover, $L^\Psi \le e^{L^\Phi}$ and $L^\Psi$ and $L^\Phi$ do not depend on $n$.
\end{lemma}

\begin{proof}:
As $V_j : \zod \rightarrow \R$ for $j \in [d]$ is Lipschitz on $\zod$, from~\cref{HongAnas} there exists a ReLU MLP $\Phi_j: \zod \rightarrow \R$ tangent to the faces of the cube and modulus of regularity $\omega_j$ with width at most $8d(n+1)^d$, depth at most $ \lceil log_2d \rceil + 4$ and at most $16d(n+1)^d$ free parameters such that:
\begin{equation}
\label{ZOD_ReuyanResult}
    \|V_j -\Phi_j\|_{L^\infty(\zod)} \leq \omega_j(\frac{d}{2n}).    
\end{equation}
If needed, one can increase the depth of the neural networks such that $\mathrm{depth}(\Phi_j) = \max_{1 \leq i \leq d} \{ \mathrm{depth}(\Phi_i) \}$, then by parallelization~\cite[Proposition 2.3.]{petersen2024mathematical} the neural network: 
\[\Phi(x)=(\Phi_1(x), \ldots, \Phi_d(x)):\mathbb{R}^{d} \rightarrow \Rd\] has the same depth. Its width is at most: 
\[
\mathrm{width}(\Phi) \leq \sum_{j=1}^d \mathrm{width}(\Phi_j) = 8d^2(n+1)^d,
\]
and free parameters at most $16d^2(n+1)^d$.
Define $( \omega)_j \eqdef  \omega_j$ then:
\begin{equation}
        \|\Phi - V\|_{ L^\infty( \zod ; l^\infty )} 
    \le
        \|\omega(\frac{d}{2n})\|_{l^\infty} ,  
\end{equation}
Note that if $V$ is such that it is tangent on the faces of the cube, then the MLP has the same value of the vector field on the vertices of boundary simplices. Also note that the proof uses hat functions and triangulation which makes the MLP tangent to the faces. This means there won't be any outward normal vector on the boundary and in particular, there will be no flow-out of $\Flow(\Phi)$.
this yields:
\begin{equation}
\label{ineq:ZOD_Upperbound_Vectorfields}
    \begin{split}
        \|V(x_s^x)-\Phi(z_s^x)\|_{L^\infty(\zod; l^\infty)} 
        &\leq 
            \|V(x_s^x)-V(z_s^x)\|_{L^\infty(\zod; l^\infty)} + \| V(z_s^x)-\Phi(z_s^x)\|_{L^\infty(\zod; l^\infty)} \\
        &\leq 
            L^{V}\| x_s^x - z_s^x \|_{L^\infty(\zod; l^\infty)} 
        + 
            \|\omega(\frac{d}{2n})\|_{l^\infty} ,  
    \end{split}
\end{equation}
hence:
\begin{equation*}
\begin{split}
    \| z_t^x - x_t^x \|_{L^\infty(\zod; l^\infty)} 
    &= 
        \| \int_0^t V( x_s^x ) - \Phi( z_s^x ) ds \|_{L^\infty(\zod; l^\infty)} \\
    &\leq 
        \int_0^t \| V( x_s^x ) - \Phi( z_s^x ) \|_{L^\infty(\zod; l^\infty)}ds \\
    &\leq 
        L^{V}\int_0^t \| x_s^x - z_s^x \|_{ L^\infty( \zod ; l^\infty) } ds + \left( \| \omega ( \frac{d}{2n} ) \|_{ l^\infty } \right) t.
\end{split}
\end{equation*}
By applying Grönwall's inequality (\cite{Pachpatte1998} Theorem 1.3.1) 
\[
    \|z_t^x-x_t^x\|_{L^\infty(\zod; l^\infty)} \leq \left( \|\omega(\frac{d}{2n})\|_{l^\infty} \right) te^{L^{V}t},
\]
let $t=1$ :
\begin{equation}
\label{ineq:ZOD_NODE_Flow}
    \| \Psi(x)-\varphi(x)\|_{L^\infty(\zod; l^\infty)} 
    = 
    \|z_1^x-x_1^x\|_{L^\infty(\zod; l^\infty)}
    \leq 
        \|\omega(\frac{d}{2n})\|_{l^\infty} e^{L^{V}}.
\end{equation}
The right hand side goes to zero as $n \rightarrow \infty$. 

Observe that the Lipschitz constant $L^{\Psi}$ does \textit{not} depend on the approximation parameter $n\in \mathbb{N}_+$. This can be shown as: 
\begin{equation}
\begin{split}
    \big\| z_t^x - z_t^y \big\|_{ L^\infty( \zod ; l^\infty )}
    &\le
        \big\| x-y \big\|_{ L^\infty( \zod ; l^\infty )} + \int_0^t \big\| \Phi(z_s^x) - \Phi(z_s^y) \big\|_{ L^\infty( \zod ; l^\infty )}ds
    \\
    &
        \le 
        \big\| x-y \big\|_{ L^\infty( \zod ; l^\infty )} + L^{\Phi}\int_0^t \big\| z_s^x - z_s^y \big\|_{ L^\infty( \zod ; l^\infty )}ds.
\end{split}
\end{equation}

By applying Grönwall's inequality \cite[Theorem 1.2.2]{Pachpatte1998} 
\[
    \|z_t^x-z_t^y\|_{L^\infty(\zod; l^\infty)} \leq \|x-y\|_{L^\infty(\zod; l^\infty)} e^{tL^{\Phi}}.
\]
Let $t=1$ :
\begin{equation}
\label{ineq:ZOD_NODE}
    \| \Psi(x)-\Psi(y)\|_{L^\infty(\zod; l^\infty)} 
    = 
    \|z_1^x-z_1^y\|_{L^\infty(\zod; l^\infty)} \leq \|x-y\|_{L^\infty(\zod; l^\infty)}e^{L^{\Phi}}.
\end{equation}
This means that $L^\Psi \leq e^{L^\Phi} $. Since $L^\Phi$ is independent of $n$, $L^\Psi$ is also independent of $n$ and the result follows.   
\end{proof}

\begin{proposition}[Universal Approximation by Deep Neural ODEs]
\label[proposition]{thrm:ZOD_Universality_of_Neural_ODEs_RH}
Let
$\varphi\in \HomeoI[\zod][I]$ and $n,d, I \in \N_+$
then, 
there exist ReLU Neural ODEs $\{\Psi_i=\mathrm{Flow}(\Phi_i)\}_{i=1}^{I}$ such that $ \Psi \eqdef \bigcirc_{i=1}^{I}\,\Psi_i$ satisfies the approximation guarantee:
\begin{equation}
\label{eq:ZOD_Universality_of_Neural_ODEs}
         \big\| \varphi - \Psi \big\|_{L^\infty(\zod; l^\infty)}
    \le
     \sum_{i=1}^{I} 
        \left( \|\omega_i(\frac{d}{2n})\|_{l^\infty}
        \prod_{j=i}^{I} e^{L^{V_j}}  \right),
\end{equation}
where $(\omega_i)_j$ is the modulus of regularity of $(V_i)_j$ for $i \in [I]$ and $j \in [d]$. The right hand side converges to zero as $n\rightarrow \infty$. \\
Moreover, $\Psi \in \Homeo[\zod]$, with Lipschitz constant at most $\prod_{i=1}^{I} e^{L^{\Phi_i}}$ and $\Phi_1,\dots,\Phi_I$ are ReLU MLP vector fields with
width at most $8d^2(n+1)^d$, depth at most $ \lceil log_2d \rceil + 4$ and at most $16d^2(n+1)^d$ free parameters.
Furthermore, $L^{\Phi_i}$ and $L^\Psi$ do not depend on the parameter $n$ for $i \in [I]$. \\ 
\end{proposition}
\begin{proof}:
    The proof is similar to~\cref{thrm:Universality_of_Neural_ODEs_RH} but instead use~\cref{lem:ZOD_Vectorfield_to_Flow}.
\end{proof}

\section{Additional Notations and Definitions for the proofs:}
\label{app:Prelim__ss:Notation}
\begin{itemize}
        \item 
            Compositional Notation:
            During the course of our analysis, it will be convenient to describe ReLU MLPs
            via the role of each of their (sets of) layers.
            Specifically, the structure of a ReLU MLP $\Phi$ is represented in the following way:
            suppose
            \(
              \Phi \;=\; \mathcal{L}_m \circ (\sigma \circ \mathcal{L}_{m-1}) \circ \cdots \circ (\sigma \circ \mathcal{L}_2) \circ (\sigma \circ \mathcal{L}_1)
            \)
            where the $\mathcal{L}_i$'s are affine transformations. We express this notationally as:
            \[
            \begin{aligned}
            \mathbf{x}
            &\Longrightarrow (\sigma\!\circ\!\mathcal{L}_{1})(\mathbf{x})
              \Longrightarrow (\sigma\!\circ\!\mathcal{L}_{2})\circ(\sigma\!\circ\!\mathcal{L}_{1})(\mathbf{x}),\\
            &\Longrightarrow (\sigma\!\circ\!\mathcal{L}_{m-1})\circ\cdots\circ(\sigma\!\circ\!\mathcal{L}_{2})\circ(\sigma\!\circ\!\mathcal{L}_{1})(\mathbf{x}),\\
            &\Longrightarrow \mathcal{L}_{m}\circ(\sigma\!\circ\!\mathcal{L}_{m-1})\circ\cdots\circ(\sigma\!\circ\!\mathcal{L}_{2})\circ(\sigma\!\circ\!\mathcal{L}_{1})(\mathbf{x}),\\
            &= \Phi(\mathbf{x})\, .
            \end{aligned}
            \]
            
            In other words, if $x_1,x_2,\ldots,x_{m-1}$ are the $1,2,\ldots,(m-1)$-th hidden layers of $\Phi$
            and $x_m$ is the output layer, then the structure of $\Phi$ is expressed as:
            \[
              x \;\Longrightarrow\; x_1 \;\Longrightarrow\; x_2 \;\Longrightarrow\; \cdots
              \;\Longrightarrow\; x_{m-1} \;\Longrightarrow\; x_m \;=\; \Phi(x)\, .
            \]
        \item 
            For a function $f: \Rd \rightarrow \Rd$ let $\mathrm{deg}(f,U,x)$ be the Brouwer degree of $f$ at the point $x \in \Rd$, assuming $x \notin f(\partial U)$ and $U$ open and bounded and $f$ is continuous on $\overline{U}$.
        \item 
            A continuum is a compact and connected metric space that contains at least two points.
        \item 
            Recurrent point:
            Let $X$ be a topological space and let $f : X \to X$ be a continuous map. 
            A point $x \in X$ is called a \emph{recurrent point} of $f$ if there exists
            a sequence of integers $(n_k)$ with $n_k \to \infty$ such that:
            \[
            f^{n_k}(x) \longrightarrow x \quad \text{as } k \to \infty.
            \]
            Equivalently, $x$ is recurrent if it belongs to its own $\omega$-limit set:
            \[
            R(f) 
            \eqdef 
                \{ y \in X : f^{n_k}(x) \to y 
                \ \text{for some sequence } n_k \to \infty \}.
            \]
        \item 
            Orbit, fixed point and periodic point:
            Let $X$ be a set and $f : X \to X$ a map. 
            For $x \in X$, the \emph{forward orbit} of $x$ under $f$ is the set:
            \[
            \mathcal{O}^+(x) \eqdef \{ f^n(x) : n \in \mathbb{N}_0 \},
            \]
            where $\mathbb{N}_0 = \{0,1,2,\dots\}$ and $f^0 = \mathrm{id}_X$.
            
            If $f$ is invertible, the \emph{(full) orbit} of $x$ is:
            \[
            \mathcal{O}(x) \eqdef \{ f^n(x) : n \in \mathbb{Z} \}.
            \] 
            A point $x \in X$ is called a fixed point of $f$ if:
            \[
            f(x) = x.
            \] 
            A point $x \in X$ is called a periodic point of period $k \geq 1$ if:
            \[
            f^k(x) = x,
            \]
            and $k$ is the smallest positive integer with this property. 
            
            The set of all periodic points of exact period $k$ is denoted:
            \[
            \operatorname{Per}_k(f) \eqdef 
            \{\, x \in X : f^k(x) = x \ \text{and} \ f^j(x) \neq x 
            \ \text{for all } 0<j<k \,\}.
            \]
\end{itemize}
Given a subset $P\subseteq M$ of a manifold $M$, for $k\in \mathbb{N}_+\cup \{\infty\}$:
\begin{itemize}
    \item 
        We say that a closed set $D \subset M$ is a flat topological closed ball if there is a surjective homeomorphism $F : \overline{B(0,1)} \rightarrow D$ that can be extended to a homeomorphism on a neighborhood of $\overline{B(0,1)}$.
    \item Define 
        \(
        F(M,K) \eqdef  \{\mathrm{Flow}(V)| V\in \mathcal{X}^\infty(M,K) \}.
        \)
        For simplicity, when there is no confusion, we might write $F(K,K)$ as $F(K)$.
\end{itemize}

\vskip 0.2in
\bibliography{Bookkeaping/refs_sorted}
\end{document}